\documentclass{article}

\usepackage{arxiv}

\usepackage[utf8]{inputenc} 
\usepackage[T1]{fontenc}    
\usepackage{hyperref}       
\usepackage{url}            
\usepackage{booktabs}       
\usepackage{amsfonts}       
\usepackage{nicefrac}       
\usepackage{microtype}      
\usepackage{lipsum}		
\usepackage{graphicx}
\usepackage{natbib}
\usepackage{doi}
\usepackage{hyperref}
\usepackage{url}
\usepackage{graphicx} 
\usepackage{booktabs}
\usepackage{tabularx}
\usepackage{array}
\usepackage{tikz}
\usetikzlibrary{positioning,fit,arrows.meta}
\usepackage{amssymb} 
\usepackage{wrapfig}
\usepackage{etoolbox}
\usepackage{caption}
\usepackage{enumitem}

\usepackage{multirow}
\usepackage[table]{xcolor}
\usepackage{amssymb}

\definecolor{flashblue}{RGB}{238,245,255}
\definecolor{traintext}{RGB}{135,135,135}
\usepackage[most]{tcolorbox}
\usepackage{ragged2e}

\usepackage[most]{tcolorbox}
\usepackage{ragged2e}

\usepackage[most]{tcolorbox}
\usepackage{ragged2e}
\setcitestyle{numbers,square,comma}
\newtcolorbox{promptbox}[1]{
  enhanced,
  breakable,
  colback=blue!2,
  colframe=blue!40!black,
  title=\textbf{#1},
  fonttitle=\small\bfseries,
  fontupper=\small\normalfont\RaggedRight,
  arc=1mm,
  boxrule=0.45pt,
  left=1.5mm,
  right=1.5mm,
  top=1.2mm,
  bottom=1.2mm,
  before skip=5pt,
  after skip=5pt
}

\newcommand{\pcode}[1]{\texttt{#1}}
\newcommand{\pkey}[1]{\textbf{#1}}

\newcommand{\displaymathsize}{\small}

\AtBeginEnvironment{equation}{\displaymathsize}
\AtBeginEnvironment{equation*}{\displaymathsize}
\AtBeginEnvironment{align}{\displaymathsize}
\AtBeginEnvironment{align*}{\displaymathsize}
\AtBeginEnvironment{gather}{\displaymathsize}
\AtBeginEnvironment{gather*}{\displaymathsize}
\AtBeginEnvironment{multline}{\displaymathsize}
\AtBeginEnvironment{multline*}{\displaymathsize}

\title{FlashBack: Knowing When to Remember in Streaming Vision-Language Models}

\date{} 					
\author{%
  \textbf{Yi Chen \quad MingMing Yu \quad Boran Wang \quad Rui-Qi Wang} \\[0.3em]
  \textbf{Xiaohang Cao \quad Chu Tang \quad Jingmin Chen \quad Jie Gu} \\[0.6em]
  \textnormal{Rightly Robotics, A4X} \\[0.3em]
  \textnormal{\texttt{ychen2@a4x.io}}
}

\renewcommand{\shorttitle}{\textit{arXiv} Template}

\begin{document}
\maketitle

\begin{abstract}
Streaming vision-language models must process continuously growing video streams
under a bounded compute budget, creating a persistent tension between real-time
perception and long-term memory. Retrieving historical information provides a
natural remedy, yet historical recall is not uniformly beneficial: unnecessary
history may introduce irrelevant context into current reasoning and interfere
with native real-time perception. Effective streaming memory should therefore
address not only what to remember, but also when and how to access it. To this
end, we introduce FlashBack, a training-free framework for selective, multi-level
memory in streaming vision-language models. Before retrieving history, FlashBack
draws on the semantic understanding of the frozen streaming VLM to infer whether
a query calls for historical evidence. This assessment determines whether
inference remains on the Native trajectory or invokes an isolated Recall
trajectory. The Recall trajectory combines recent context with retrieved
long-term memory through a query-local Side-KV pathway, preserving local temporal
continuity without modifying the persistent Native state. We instantiate FlashBack on StreamingVLM and Mage-VL-4B and evaluate it
on OVO-Bench and StreamingBench.
The results show improvements on several long-horizon and memory-dependent tasks
while largely preserving real-time perception, with performance competitive with
strong training-based streaming methods despite requiring no additional training. Our code will be announced later.
\end{abstract}

\section{Introduction}
\begin{figure}[h]
    \vspace{-12pt}
    \centering
    \includegraphics[width=0.9\linewidth]{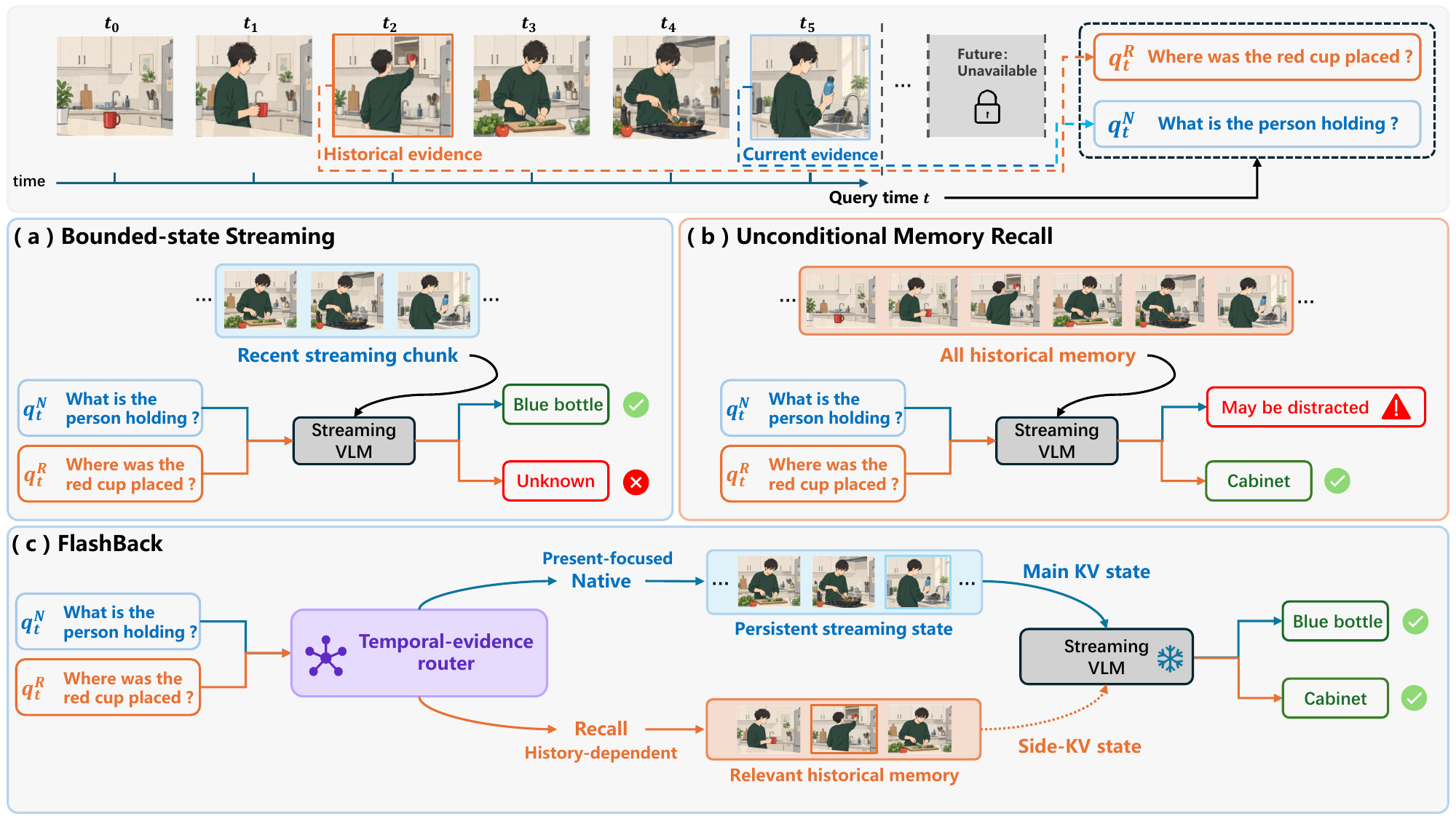}
    \caption{Motivation for FlashBack.}
    \label{fig:intro}
\end{figure}

Vision-language models (VLMs) can describe scenes and answer questions about images and videos. In offline settings, the model can examine a recorded video before answering questions about it \citep{fu2025video}. Live commentary and visual assistance, however, require the model to keep watching as new frames arrive and answer using only what it has seen so far \citep{chen2024videollm}. As shown in Figure~\ref{fig:intro}, questions asked at the same moment can refer to what is happening now or to events from much earlier in the video \citep{niu2025ovo}. The model must therefore keep up with the stream using limited computation and working memory, while still being able to answer questions about the past \citep{xu2026streamingvlm}.

Streaming models can limit computation and memory use by keeping only recent video content in the input or cache \citep{xu2026streamingvlm,shen2026simple}. This supports questions about the current scene, but earlier details may be lost: in Figure~\ref{fig:intro}(a), the model can identify the blue bottle being held but cannot recall where the red cup was placed. Memory-augmented methods keep information from earlier in the video so that it remains available for later questions \citep{zhang2025flash,di2025streaming}. This can help recover past events, but adding history can also make answers about the current scene less accurate \citep{shen2026simple}, as illustrated in Figure~\ref{fig:intro}(b).

Existing streaming memory systems can be understood along two complementary stages: how history is stored and how it is accessed. During streaming, past observations may be compressed into bounded visual or KV states, organized into higher-level scene, event, or semantic memories, or maintained as structured records that support later reasoning \citep{zhang2025flash,zhang2026hermes,liang2026oasis,fan2026folio,long2026seeing,wen2026eventmemagent}. When a question arrives, the maintained history may be used directly, selectively retrieved according to the query, adaptively expanded according to the question or model uncertainty, or searched through multi-step agentic reasoning and tool use \citep{di2025streaming,ning2025livevlm,liang2026oasis,wu2026semantic,ge2026should,long2026seeing,wen2026eventmemagent,li2026visual}. These mechanisms are often combined: a stream may first be compressed or structured into memory and then selectively searched when evidence from the past is needed. Recent work has therefore moved beyond simply retaining more history toward controlling what historical evidence is exposed to the model, how much is retrieved, and how that evidence is gathered for answering. For a model that already supports native streaming inference, however, there remains a distinct decision before these forms of memory access become necessary: whether the temporal evidence requested by the current question calls for historical memory in the first place. This motivates treating memory use as a choice between the Native and Recall trajectories before retrieval. A streaming model therefore needs to decide not only what history to keep and how to retrieve it, but when history should participate in answering.

Selecting which historical evidence to retrieve and deciding whether to recall are not the same problem. The former determines what information should be brought into memory-augmented reasoning, whereas the latter determines whether the model should enter such a reasoning path at all. This distinction motivates us to treat memory use as a choice between answer trajectories rather than only as evidence selection within a memory-augmented path. We introduce FlashBack, a training-free selective-memory framework built on frozen streaming VLMs. Before retrieving history, FlashBack uses the semantic understanding provided by the frozen VLM to infer the temporal-evidence
requirement of the query and maps this requirement to either the Native or
Recall trajectory. Queries assigned to Native preserve the original streaming
inference behavior, whereas Recall queries access historical evidence together
with recent context through a query-local Side-KV pathway. The retrieved history
and recent context are processed using standard causal attention in the frozen
decoder, and the resulting temporary KV states are discarded after answering
rather than written back to the persistent Native trajectory. FlashBack therefore
makes memory augmentation conditional on the query, introducing historical
evidence when needed while leaving the Native streaming trajectory unchanged
otherwise. Our contributions are as follows:
\begin{itemize}[leftmargin=*,nosep]
\setlength{\itemsep}{4pt}
\setlength{\parskip}{0pt}
\setlength{\parsep}{1pt}
\setlength{\topsep}{2pt}
\item We introduce \textbf{FlashBack}, a training-free selective-memory framework that uses the temporal-evidence understanding of the frozen VLM to decide when historical recall is needed. Across two streaming VLMs on OVO-Bench and StreamingBench, FlashBack improves
several long-horizon and memory-dependent tasks while largely preserving
real-time perception.
\item We introduce Side-KV, an isolated query-local memory-access pathway that incorporates recalled history without modifying the Native streaming state.

\item We develop a multi-level Recall design that combines recent context with query-relevant long-term memory to support both local temporal continuity and distant recall.
\end{itemize}

\section{Related Work}
\noindent\textbf{Streaming Vision-Language Models}\quad Early approaches to online video understanding largely obtain streaming capability by adapting VLMs whose core perception and reasoning pipelines are designed around finite images or video clips. At the simplest end, an off-the-shelf VLM can process only a sliding window of recent frames \citep{shen2026simple}; more elaborate systems introduce online memory, streaming-oriented interaction, or other mechanisms around conventional video-language backbones \citep{chen2024videollm,huang2025online}. These approaches make continuous inputs tractable, but streaming is largely enabled by the surrounding inference or adaptation procedure rather than by persistent causal state reuse within the model itself.

More recent \emph{native streaming} models instead maintain and reuse online state as the input stream grows. StreamingVLM keeps a bounded persistent KV state composed of attention sinks, recent visual tokens, and a longer text context, enabling causal processing over effectively unbounded video streams \citep{xu2026streamingvlm}. Mage-VL-4B extends streaming processing to the visual front end through codec-aware incremental encoding, using motion and residual information to retain informative visual patches together with a dual-system architecture for proactive interaction \citep{yang2026mage}. We instantiate FlashBack on StreamingVLM and Mage-VL-4B, two streaming VLMs with different online processing designs, to evaluate FlashBack beyond a single streaming architecture.

\noindent\textbf{Memory for Streaming Video Understanding}\quad Memory mechanisms extend streaming VLMs beyond the limited information retained in their active context. Earlier approaches compress past visual information or manage KV caches to preserve useful history under a constrained budget, while retrieval-based methods store historical representations separately and recover query-relevant units when needed \citep{zhang2025flash,zhang2026hermes,di2025streaming,ning2025livevlm,azad2026streamready,hao2026shallowstream}. Other systems organize history into higher-level episodic, event, or semantic memories. More recently, agentic approaches further turn memory access into an active process, using multi-step reasoning, memory search, or external perception tools to gather evidence from long streams \citep{long2026seeing,wen2026eventmemagent,li2026visual}. These designs are complementary: compression and structuring determine how history is maintained, while retrieval and agentic search determine how stored evidence is brought back into the reasoning process.

Recent work has increasingly made memory access selective. OASIS expands from short-context reasoning to hierarchical event retrieval when additional evidence is needed, SAVEMem adapts retrieval scope according to the temporal requirements of a query, and StreamRAG dynamically adjusts retrieval using temporal sensitivity and contextual relevance \citep{liang2026oasis,wu2026semantic,xie2026streamrag}. SelectStream further learns when to write, what to preserve, and how to retrieve historical evidence under a fixed latent-memory budget \citep{ge2026should}, while FOLIO combines short-term visual context with structured semantic memory and query-dependent evidence assembly without additional training \citep{fan2026folio}. These methods substantially improve how historical evidence is retained,
retrieved, or assembled once memory is involved in reasoning. FlashBack
focuses on a complementary decision level: for a native streaming model, it
first determines whether memory-augmented reasoning should be invoked at all. The temporal-evidence requirement of each query is mapped to either the Native or Recall trajectory before retrieval, and only Recall accesses long-term memory through Side-KV. History is therefore treated as a conditional source of evidence rather than the default context for answering.

\section{Method}
\subsection{Overview}
\begin{figure}[h]
    \centering
    \includegraphics[width=0.95\linewidth]{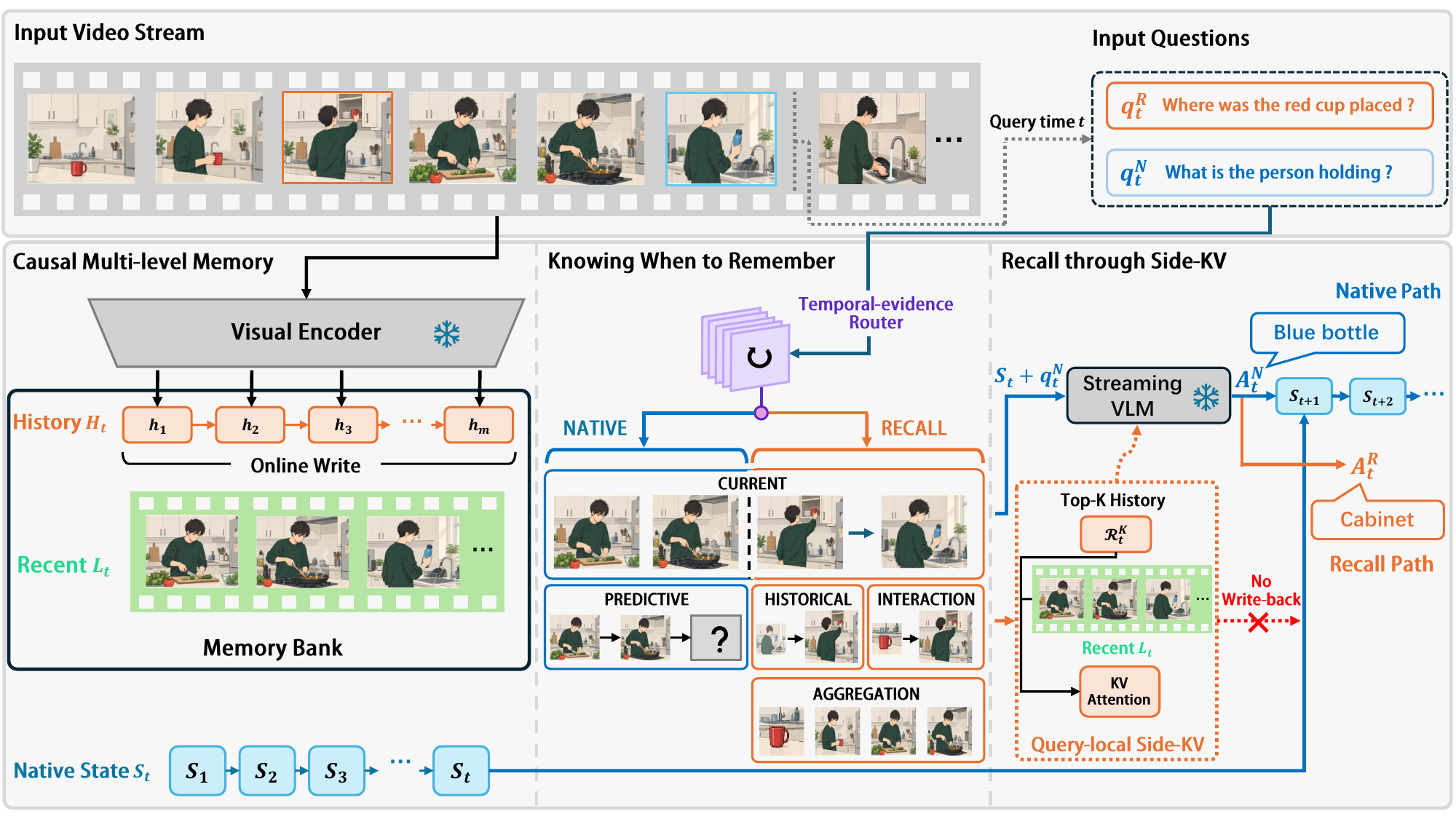}
    \vspace{-3pt}
    \caption{Overview of FlashBack.}
    \label{fig:method}
    \vspace{-5mm}
\end{figure}

Figure~\ref{fig:method} presents the overall architecture of FlashBack. Built on a frozen streaming VLM, FlashBack preserves the original Native streaming trajectory while maintaining recent context and causal long-term memory as the video progresses. When a query arrives, a temporal-evidence router determines whether the answer should remain on the Native trajectory or invoke Recall. Recall combines recent context with retrieved long-term memory through a query-local Side-KV pathway, without modifying the persistent Native state. We next describe the organization of these in detail.

\subsection{Causal Multi-level Memory}
FlashBack maintains a causal memory bank alongside the persistent Native streaming trajectory. At time \(t\), the bank is built only from the observed video prefix \(X_{\leq t}\): no future frames are available, and memory is written online as the stream progresses rather than reconstructed from the full video after a query arrives. As illustrated in Figure~\ref{fig:method}, we organize the bank into two temporal levels,
\begin{equation}
    B_t = \left(H_t, L_t\right),
\end{equation}
where \(L_t\) covers the recent visual context near the current streaming boundary and \(H_t\) stores completed historical units from earlier parts of the stream. The bank is maintained alongside the Native trajectory, so historical evidence remains available without becoming part of the active reasoning context by default.
The recent level retains the backbone-native visual states within a moving temporal window,
\begin{equation}
    L_t = \{r_j\}_{j \in \mathcal{W}_t},
\end{equation}
where \(\mathcal{W}_t\) denotes the recent range at time \(t\). These states preserve local temporal continuity around the query boundary, including recent events and state changes that may be difficult to recover from an isolated historical unit. The exact form of \(r_j\) follows the underlying streaming backbone, allowing FlashBack to reuse the visual states already produced during online processing rather than imposing a shared cache representation across different models.
The long-term level contains temporally bounded historical units,
\begin{equation}
    H_t = \{h_i\}_{i=1}^{N_t},
    \qquad
    h_i = \left(R_i, \Delta_i\right),
\end{equation}
where \(R_i\) contains the model-internal visual representation and cache information required to later access the \(i\)-th unit, and $\Delta_i = [t_i^{\mathrm{s}}, t_i^{\mathrm{e}}]$ records the corresponding interval in the video stream. The physical form of \(R_i\) is backbone-dependent, but the stored content remains in the internal visual representation space rather than being converted into textual summaries. The interval \(\Delta_i\) serves as temporal metadata for causal organization
and later memory access. A unit is added to \(H_t\) only after it has been
fully observed. As the stream advances, \(L_t\) moves with the current
boundary, while completed earlier units are progressively retained in \(H_t\).

The two levels therefore keep complementary temporal evidence available for later Recall. \(L_t\) preserves continuity close to the current streaming boundary, whereas \(H_t\) retains evidence from more distant observations. Together, they form a causal multi-level memory bank that is constructed online before future queries are known.

\subsection{Knowing When to Remember}

Constructing a memory bank determines what temporal evidence remains available, but not whether that evidence should participate in a particular answer. FlashBack therefore does not treat memory access as a direct binary gate. Instead, the Temporal-evidence router first interprets the temporal support required by a query and then grounds this semantic state to an inference action:
\begin{equation}
    c_t = g_{\theta}(q_t),
    \qquad
    a_t = \phi(c_t) \in \{\mathrm{Native},\mathrm{Recall}\}.
\end{equation}
Here, \(g_{\theta}\) uses the frozen VLM to infer temporal-evidence semantics, while \(\phi\) is a fixed semantic-to-action mapping. These semantics are not benchmark task labels, retrieval scopes, or templates for assembling memory evidence. They determine whether inference should remain on the Native trajectory or invoke Recall. Importantly, this assessment is completed before accessing \(H_t\), so retrieval follows the Recall decision rather than informing it.
The first stage operates over a general temporal-evidence space,
\begin{equation}
\begin{aligned}
\mathcal{C}^{(1)}
&=
\{\mathrm{Current},\mathrm{Historical},\mathrm{Interaction},
  \mathrm{Aggregation},\mathrm{Predictive}\},\\
\phi^{(1)} &:\
\{\mathrm{Historical},\mathrm{Interaction},\mathrm{Aggregation}\}
\mapsto \mathrm{Recall},\\
&\qquad
\mathrm{Predictive}\mapsto\mathrm{Native},\quad
\mathrm{Current}\mapsto\mathrm{Refine}.
\end{aligned}
\end{equation}
The five states correspond to current perception, an earlier observed state or event, temporal relations among observed events, evidence distributed over the observed prefix, and an outcome that has not yet been observed. Current perception is refined because reference to the present does not necessarily mean that the current visual boundary alone provides sufficient evidence.
For Current queries, FlashBack performs a second semantic assessment,
\begin{equation}
\begin{aligned}
\mathcal{C}^{(2)}
&=
\{\mathrm{Boundary},\mathrm{Ongoing},\mathrm{Persistent},\mathrm{Outcome}\},\\
\phi^{(2)} &:\
\{\mathrm{Boundary},\mathrm{Ongoing}\}\mapsto\mathrm{Native},\qquad
\{\mathrm{Persistent},\mathrm{Outcome}\}\mapsto\mathrm{Recall}.
\end{aligned}
\end{equation}
Here, Boundary and Ongoing describe directly observable configurations or processes still unfolding at the query boundary, whereas Persistent and Outcome require evidence established by earlier observations. This refinement makes the decision depend on temporal support rather than surface cues such as ``now'' or ``before''.

To obtain these semantic decisions without additional training, FlashBack performs representation-balanced semantic inference. A single categorical prompt can be sensitive to category order and arbitrary output codes. For each stage \(k\), we construct semantically equivalent views \(\mathcal{V}^{(k)}\) by jointly permuting category order and category-to-code assignment. Let \(P_v^{(k)}(q_t)\) denote view \(v\), and \(z_v^{(k)}(c)\) the code assigned to semantic state \(c\). We aggregate the log probabilities after mapping them back to the shared semantic space:
\vspace{-2mm}
\begin{equation}
\begin{aligned}
\bar{s}^{(k)}(c\mid q_t)
&=
\frac{1}{|\mathcal{V}^{(k)}|}
\sum_{v\in\mathcal{V}^{(k)}}
\log p_{\theta}
\left(
z_v^{(k)}(c)\mid P_v^{(k)}(q_t)
\right),\\
\hat{c}_t^{(k)}
&=
\arg\max_{c\in\mathcal{C}^{(k)}}
\bar{s}^{(k)}(c\mid q_t).
\end{aligned}
\end{equation}
The first stage uses five balanced views, while queries assigned to Current enter a second stage with four views. These views are equivalent parameterizations of the same semantic decision, and their mapped log probabilities are averaged rather than combined by majority voting. The final action is
\begin{equation}
a_t =
\begin{cases}
\phi^{(1)}(\hat{c}_t^{(1)}),
    & \hat{c}_t^{(1)} \neq \mathrm{Current},\\
\phi^{(2)}(\hat{c}_t^{(2)}),
    & \hat{c}_t^{(1)} = \mathrm{Current}.
\end{cases}
\end{equation}
As shown in Figure~\ref{fig:method}, Native queries form \(q_t^N\) and remain on the persistent streaming trajectory, whereas Recall queries form \(q_t^R\). Only \(q_t^R\) triggers retrieval from \(H_t\) and the construction of Side-KV. The Temporal-evidence router therefore separates \emph{whether} historical evidence is required from the subsequent problem of \emph{which} historical evidence should be retrieved. The complete routing prompts and representation-balanced view construction are provided in Appendix~\ref{app:router}.

\subsection{Recall through Side-KV}

Once a query is routed to Recall, FlashBack must determine which historical evidence to retrieve and how that evidence should enter inference. Retrieval is performed over the causal memory bank, while the resulting Recall computation remains separate from the persistent Native trajectory. This distinction is important for streaming models, where hidden or KV states are continuously reused by subsequent inputs \citep{dai2019transformer,xiao2024efficient}. Historical evidence selected for a particular query should support the current answer without becoming part of the state inherited by future observations.

\noindent\textbf{Causal long-term retrieval.}
Only \(q_t^R\) triggers access to the long-term history \(H_t\). Let \(t_L^{\mathrm{s}}\) denote the beginning of the recent range represented by \(L_t\). Using the temporal interval associated with each historical unit, we restrict retrieval to
\begin{equation}
    \mathcal{E}_t =
    \left\{
        h_i=(R_i,\Delta_i)\in H_t
        \;\middle|\;
        t_i^{\mathrm{e}} < t_L^{\mathrm{s}}
    \right\}.
\end{equation}
This restriction uses only completed historical units preceding the recent
context. A frozen, question-conditioned prefilter first forms
\(\mathcal{C}_t\subseteq\mathcal{E}_t\), after which FlashBack reranks the
candidates using model-internal query--key compatibility and retrieves the top
\(K\),
\begin{equation}
\mathcal{R}_t^{K}
=
\operatorname{TopK}_{h_i\in\mathcal{C}_t}
s_{\theta}(q_t^R,L_t,h_i).
\end{equation}
where \(\mathcal{R}_t^{K}\) denotes the selected long-term units. Retrieval
remains causal and uses neither future observations nor answer correctness.
Implementation details are provided in Appendix~\ref{app:retrieval}.

\noindent\textbf{Multi-level Recall.}
Let \(R(\mathcal{R}_t^{K})\) denote the internal representations of the retrieved long-term units, preserving their temporal organization. FlashBack combines them with the recent level of the memory bank:
\begin{equation}
    C_t^{R} =
    \begin{cases}
        [R(\mathcal{R}_t^{K});L_t], & \mathcal{E}_t \neq \varnothing,\\
        L_t, & \mathcal{E}_t = \varnothing.
    \end{cases}
\end{equation}
The second case remains a recent-only Recall rather than returning to Native. This combination provides access to distant historical evidence while preserving local temporal continuity near the current streaming boundary.

\noindent\textbf{Query-local Side-KV.}
Recall is executed by the same frozen causal decoder, without an additional fusion network or learned memory reader. We denote the temporary KV trajectory constructed from the Recall context as
\begin{equation}
    \mathrm{SideKV}_t
    =
    \operatorname{KV}_{\theta}
    \left(
        [C_t^{R};q_t^R]
    \right),
\end{equation}
where \(\operatorname{KV}_{\theta}\) denotes the query-local KV states
constructed by the frozen decoder under the backbone-specific cache layout.
When long-term evidence is available, retrieved historical and recent states
are jointly exposed to the query within the Recall trajectory; otherwise,
Side-KV uses the recent context alone.

Writing query-specific recalled states back into the persistent cache would cause later observations to be processed under an attention context augmented with evidence selected for an earlier query. Subsequent hidden states and KV states would then be generated under this retrieval-conditioned context, allowing the effect of a single Recall operation to propagate along the future Native trajectory. Since future queries are unknown during streaming, such query-specific retrieval should remain local to the query that requested it \citep{kim2025infinipotv}. FlashBack therefore adopts No Write-back; Side-KV is discarded after the current answer, allowing recalled evidence to support the present inference without altering the persistent Native trajectory used by subsequent observations and queries. Implementation
details are provided in Appendix~\ref{app:sidekv}.

\section{Experiment}
\subsection{Implementation Details}
All experiments are conducted using \(4\times\) NVIDIA A100 GPUs (80GB). We implement FlashBack on StreamingVLM \citep{xu2026streamingvlm} and Mage-VL-4B \citep{yang2026mage} without additional training. Ablation studies are conducted on StreamingVLM, while the main experiments use both backbones.

For finite-answer tasks, we follow the candidate-logit evaluation protocol used by StreamingVLM: the allowed answer candidates are scored using next-token logits after a fixed answer prefix, and the candidate with the highest logit is selected. This reduces variation arising from free-form instruction following and
answer formatting. Detailed settings are provided in Appendix ~\ref{app:main setting} and ~\ref{app:scoring}.

\subsection{Datasets}
\label{sec:datasets}
We evaluate FlashBack on OVO-Bench \citep{niu2025ovo} and
StreamingBench \citep{lin2026streamingbench}. Component ablations use
StreamingBench-MA for cue-to-scene alignment and OVO-Bench-CRR for determining
when sufficient evidence has accumulated. Together, they probe selective
history use and temporal evidence accumulation.

\subsection{Ablation Study}

\leavevmode\par
\vspace{-6pt}

\begingroup
\setlength{\intextsep}{1pt}
\setlength{\abovecaptionskip}{0pt}
\setlength{\belowcaptionskip}{2pt}
\setlength{\columnsep}{8pt}

\begin{wraptable}{r}{0.44\columnwidth}
\vspace{-30pt}
\centering
\captionsetup{
    font=footnotesize,
    labelfont=normalfont,
    textfont=bf,
    skip=1pt
}
\caption{Effect of Recall and Routing.}
\label{tab:ablation-routing}
\vspace{-0.5mm}

\footnotesize
\setlength{\tabcolsep}{3.0pt}
\renewcommand{\arraystretch}{1.04}

\begin{tabular*}{\linewidth}{@{\extracolsep{\fill}}lcc@{}}
\toprule
\textbf{Method} & \textbf{MA} & \textbf{CRR} \\
\midrule
Native
& 69.60 & 41.67 \\
+ Recall
& 70.40 & \textbf{60.83} \\
+ Direct Routing
& 69.20 & 41.67 \\
+ Structured Routing
& \textbf{72.80} & \textbf{60.83} \\
\bottomrule
\end{tabular*}
\end{wraptable}

\noindent\textbf{Recall and Routing.}
Table~\ref{tab:ablation-routing} separates the contribution of historical
Recall from the decision of when to invoke it. Recall produces a large gain on
CRR, where answering depends on evidence accumulated over the stream, while
changing MA only modestly. Direct binary routing loses the CRR improvement and
returns performance close to the Native baseline. In contrast, structured
routing preserves the Recall gain on CRR and further improves MA. These results
suggest that explicitly identifying the temporal-evidence requirement provides
a more reliable basis for choosing between the Native and Recall trajectories
than predicting the answer branch directly. Detailed analysis is provided in
Appendix~\ref{app:trajectory-selection}.

\par

\begin{wraptable}{r}{0.44\columnwidth}
\vspace{-6pt}
\centering
\captionsetup{
    font=footnotesize,
    labelfont=normalfont,
    textfont=bf,
    skip=1pt
}
\caption{Effect of Memory Hierarchy.}
\label{tab:ablation-memory}
\vspace{-0.5mm}

\footnotesize
\setlength{\tabcolsep}{3.0pt}
\renewcommand{\arraystretch}{1.04}

\begin{tabular*}{\linewidth}{@{\extracolsep{\fill}}lcc@{}}
\toprule
\textbf{Memory} & \textbf{MA} & \textbf{CRR} \\
\midrule
Recent only (\(K=0\))
& \textbf{73.60} & 55.42 \\
\(K=1\)
& 72.80 & \textbf{60.83} \\
\(K=3\)
& 72.80 & \textbf{60.83} \\
\(K=5\)
& 72.40 & \textbf{60.83} \\
\bottomrule
\end{tabular*}

\vspace{3pt}
\end{wraptable}

\noindent\textbf{Memory Hierarchy.}
Table~\ref{tab:ablation-memory} compares recent-only Recall with access to
long-term historical pages. Adding one historical page improves CRR, where
answering depends on evidence accumulated beyond the recent context, while MA
slightly favors the recent-only setting. Increasing the retrieval budget beyond
\(K=1\) provides no further gain on CRR and slightly reduces MA at \(K=5\).
We therefore use \(K=1\) as the default, providing access to long-term evidence
with a compact Recall context.

\par
\endgroup

\begingroup
\setlength{\intextsep}{0pt}
\setlength{\abovecaptionskip}{0pt}
\setlength{\belowcaptionskip}{2pt}
\setlength{\columnsep}{8pt}

\noindent\textbf{Side-KV.}
Table~\ref{tab:ablation-sidekv} isolates the memory interface with routing
disabled, applying each integration strategy to every query. All variants use the same settings except for the memory integration method. The first three variants append the complete historical page to a
query-specific clone of the Native cache for the answer forward, whereas
Side-KV exposes the same page through a query-local Recall trajectory.
Historical KV substantially outperforms the random and zero controls on both
subsets, showing that the retrieved page contains task-relevant information.

\begin{wraptable}{r}{0.40\columnwidth}
\vspace{-10pt}
\centering
\captionsetup{
    font=footnotesize,
    labelfont=normalfont,
    textfont=bf,
    skip=1pt
}
\caption{Effect of Side-KV.}
\label{tab:ablation-sidekv}
\vspace{-0.5mm}

\footnotesize
\setlength{\tabcolsep}{3.0pt}
\renewcommand{\arraystretch}{1.04}

\begin{tabular*}{\linewidth}{@{\extracolsep{\fill}}lcc@{}}
\toprule
\textbf{Integration} & \textbf{MA} & \textbf{CRR} \\
\midrule
Random KV
& 33.20 & 39.58 \\
Zero KV
& 26.80 & 39.17 \\
Historical KV
& 64.40 & 45.83 \\
\midrule
Side-KV
& \textbf{70.40} & \textbf{60.83} \\
\bottomrule
\end{tabular*}

\vspace{-1pt}
\end{wraptable}
Side-KV further improves both MA and CRR, with a particularly large margin on
CRR. Since the selected historical page is shared across the compared
interfaces, the results indicate that performance depends not only on what
history is retrieved, but also on how that evidence is incorporated into
streaming inference. 
The CRR result further suggests that query-local access is
especially useful when evidence accumulated over the stream must be integrated
at answer time. 
Detailed implementation is provided in Appendix~\ref{app:sidekv}.
\vspace{-5pt}
\par
\endgroup

\subsection{Main Results}
\leavevmode\par
\vspace{-6pt}
\setlength{\columnsep}{8pt}
\setlength{\intextsep}{0pt}

\begin{wraptable}{r}{0.58\columnwidth}
\vspace{-36pt}
\centering
\captionsetup{
    font=footnotesize,
    labelfont=normalfont,
    textfont=bf,
    skip=3pt
}
\caption{OVO-Bench group-level results.}
\label{tab:ovo-comparison}
\vspace{-1mm}

\footnotesize
\renewcommand{\arraystretch}{1.04}

\resizebox{\linewidth}{!}{%
\begin{tabular}{@{}lccc@{}}
\toprule
\textbf{Method}
& \textbf{RT Avg.}
& \textbf{Bwd. Avg.}
& \textbf{Fwd. Avg.} \\
\midrule

\multicolumn{4}{@{}l}{
    \color{traintext}\textit{Training-based methods}
} \\

VideoLLM-online
& 20.79 & 17.73 & -- \\

EventMemAgent-8B
& 68.29 & 58.03 & 55.92 \\

SelectStream-Qwen3-8B
& \textbf{82.76} & 62.20 & 56.13 \\

StreamReady-7B
& 73.60 & \textbf{72.20} & \textbf{58.80} \\

\midrule

\multicolumn{4}{@{}l}{
    \color{traintext}\textit{Training-free methods}
} \\

StreamRAG-ViSpeak-7B
& 69.68 & 60.78 & 54.80 \\

HERMES-7B
& 68.98 & 49.43 & -- \\

SimpleStream-Qwen3-8B
& 81.40 & 54.00 & -- \\

OASIS-Qwen3-8B
& 78.14 & 57.21 & -- \\

SAVEMem-Qwen2.5-7B
& 74.93 & 50.44 & -- \\

FOLIO-Qwen3-8B
& 82.00 & 69.10 & -- \\

\midrule

\multicolumn{4}{@{}l}{
    \color{traintext}\textit{Ours (training-free)}
} \\

StreamingVLM-7B
& 71.66 & 46.53 & 34.59 \\

\rowcolor{flashblue}
\textbf{StreamingVLM + FlashBack-7B}
& 70.85 & 44.92 & 41.22 \\

Mage-VL-4B
& 81.78 & 47.86 & 45.17 \\

\rowcolor{flashblue}
\textbf{Mage-VL + FlashBack-4B}
& 81.20 & 49.93 & 51.94 \\

\bottomrule
\end{tabular}%
}

\vspace{-5pt}
\end{wraptable}

\noindent\textbf{OVO-Bench.}
Table~\ref{tab:ovo-comparison} shows that FlashBack provides its clearest
group-level improvement on Forward reasoning while keeping Realtime performance
close to the corresponding Native trajectories. Table~\ref{tab:ovo-detailed}
further highlights representative task-level effects according to their actual
evidence requirements rather than their top-level benchmark categories.

EPM, ASI, REC, and CRR capture different forms of temporal dependence. EPM
requires recalling a previously observed event, ASI requires preserving the
order of completed actions, REC relates the current event to earlier
repetitions, and CRR accumulates clues before answering. FlashBack yields its
most consistent gains on REC and CRR, while improvements on EPM and ASI are
more backbone dependent. In particular, EPM remains unchanged on StreamingVLM
and improves slightly on Mage-VL, indicating that explicit access to historical
memory does not necessarily translate into uniform gains across all episodic
queries.
\begingroup
\setlength{\intextsep}{0pt}
\setlength{\abovecaptionskip}{0pt}
\setlength{\belowcaptionskip}{2pt}
\setlength{\columnsep}{8pt}

\begin{wraptable}{r}{0.51\columnwidth}
\vspace{-8pt}
\centering
\captionsetup{
    font=footnotesize,
    labelfont=normalfont,
    textfont=bf,
    skip=2pt
}
\caption{Memory-related tasks on OVO-Bench.}
\label{tab:ovo-detailed}
\vspace{-0.5mm}

\footnotesize
\setlength{\tabcolsep}{2.4pt}
\renewcommand{\arraystretch}{1.04}

\resizebox{\linewidth}{!}{%
\begin{tabular}{@{}l|ccccc@{}}
\toprule
\textbf{Method}
& \textbf{EPM}
& \textbf{ASI}
& \textbf{REC}
& \textbf{CRR}
& \textbf{HLD} \\
\midrule

StreamingVLM
& \textbf{54.88}
& \textbf{56.76}
& 19.34
& 41.67
& \textbf{27.96} \\

\rowcolor{flashblue}
\textbf{StreamingVLM + FlashBack}
& \textbf{54.88}
& \textbf{56.76}
& \textbf{20.06}
& \textbf{60.83}
& 23.12 \\

\midrule

Mage-VL
& 52.53
& 58.78
& 19.48
& 44.17
& \textbf{32.26} \\

\rowcolor{flashblue}
\textbf{Mage-VL + FlashBack}
& \textbf{53.20}
& \textbf{64.86}
& \textbf{29.80}
& \textbf{54.17}
& 31.72 \\

\bottomrule
\end{tabular}%
}

\vspace{-2pt}
\end{wraptable}

HLD exhibits a different trade-off. Unlike the other selected tasks, which
require positively observed historical evidence, HLD evaluates whether the
model rejects a question that is unsupported by the video. FlashBack reduces
HLD accuracy on both backbones, with a larger decrease on StreamingVLM. This
suggests that retrieved history can be temporally or semantically related to a
query without constituting valid evidence for its premise, making unsupported
questions harder to reject. The contrast with EPM, ASI, REC, and CRR reinforces
the query-dependent role of memory: historical access is beneficial when it
provides evidence required by the query, but additional history is not uniformly
helpful for every form of temporal reasoning.
\vspace{6pt}
\par
\WFclear
\endgroup

\begingroup
\setlength{\intextsep}{0pt}
\setlength{\abovecaptionskip}{0pt}
\setlength{\belowcaptionskip}{2pt}
\setlength{\columnsep}{8pt}

\begin{wraptable}{r}{0.54\columnwidth}
\vspace{-20pt}
\centering
\captionsetup{
    font=footnotesize,
    labelfont=normalfont,
    textfont=bf,
    skip=2pt
}
\caption{StreamingBench group-level results.}
\label{tab:main-streamingbench}
\vspace{-0.5mm}

\footnotesize
\setlength{\tabcolsep}{2.8pt}
\renewcommand{\arraystretch}{1.04}

\resizebox{\linewidth}{!}{%
\begin{tabular}{@{}lccc@{}}
\toprule
\textbf{Method}
& \textbf{RT Avg.}
& \textbf{Omni Avg.}
& \textbf{Context Avg.} \\
\midrule

\multicolumn{4}{@{}l}{
    \color{traintext}\textit{Training-based methods}
} \\

VideoLLM-online
& 35.99 & 28.45 & -- \\

EventMemAgent-8B
& 77.00 & -- & -- \\

SelectStream-Qwen3-8B
& \textbf{82.67} & -- & -- \\

\midrule

\multicolumn{4}{@{}l}{
    \color{traintext}\textit{Training-free methods}
} \\

StreamRAG-ViSpeak-7B
& 78.12 & -- & -- \\

HERMES-7B
& 79.44 & -- & -- \\

SimpleStream-Qwen3-8B
& 80.59 & -- & -- \\

\midrule

\multicolumn{4}{@{}l}{
    \color{traintext}\textit{Ours (training-free)}
} \\

StreamingVLM Native-7B
& 78.82 & 52.90 & 48.13 \\

\rowcolor{flashblue}
\textbf{StreamingVLM + FlashBack-7B}
& 79.22 & 53.30 & 48.27 \\

Mage-VL Native-4B
& 81.63 & \textbf{65.20} & \textbf{62.13} \\

\rowcolor{flashblue}
\textbf{Mage-VL + FlashBack-4B}
& 82.59 & 65.10 & 61.07 \\

\bottomrule
\end{tabular}%
}

\vspace{-4pt}
\end{wraptable}

\noindent\textbf{StreamingBench.}
Table~\ref{tab:main-streamingbench} reports performance across the three
StreamingBench task groups. FlashBack improves Real-Time Visual Understanding
on both backbones, with Mage-VL + FlashBack approaching the strongest
training-based result. StreamingVLM also improves slightly on Omni-Source and
Contextual Understanding. Mage-VL shows a slight decrease on Omni-Source and a
larger reduction on Contextual Understanding, indicating that the effect of
historical access is not uniform across all forms of streaming context.
The group averages nevertheless combine tasks with substantially different
temporal-evidence requirements. Some Real-Time tasks require information to be
maintained or accumulated over the stream, whereas some Contextual tasks require
the model to suppress historical information that is no longer relevant.
Task-level results therefore provide a more direct view of where memory access
is beneficial.

\par
\WFclear
\endgroup

\begingroup
\setlength{\intextsep}{0pt}
\setlength{\abovecaptionskip}{0pt}
\setlength{\belowcaptionskip}{2pt}
\setlength{\columnsep}{8pt}

\begin{wraptable}{r}{0.54\columnwidth}
\vspace{-6pt}
\centering
\captionsetup{
    font=footnotesize,
    labelfont=normalfont,
    textfont=bf,
    skip=2pt
}
\caption{Memory-related tasks on StreamingBench.}
\label{tab:streamingbench-memory}
\vspace{-0.5mm}

\footnotesize
\setlength{\tabcolsep}{2.6pt}
\renewcommand{\arraystretch}{1.04}

\resizebox{\linewidth}{!}{%
\begin{tabular}{@{}l|ccccc@{}}
\toprule
\textbf{Method}
& \textbf{CR}
& \textbf{CS}
& \textbf{EU}
& \textbf{CT}
& \textbf{ACU} \\
\midrule

StreamingVLM
& 79.69
& 87.70
& \textbf{80.38}
& 27.66
& 52.40 \\

\rowcolor{flashblue}
\textbf{StreamingVLM + FlashBack}
& \textbf{81.25}
& \textbf{88.33}
& \textbf{80.38}
& \textbf{30.32}
& \textbf{53.60} \\

\midrule

Mage-VL
& 67.97
& \textbf{90.22}
& 76.58
& 40.43
& \textbf{67.20} \\

\rowcolor{flashblue}
\textbf{Mage-VL + FlashBack}
& \textbf{71.88}
& \textbf{90.22}
& \textbf{81.01}
& \textbf{46.28}
& \textbf{67.20} \\

\bottomrule
\end{tabular}%
}

\vspace{-2pt}
\end{wraptable}

Table~\ref{tab:streamingbench-memory} highlights tasks that require temporal
context, historical retention, or evidence accumulation and do not decrease on
either backbone. CR requires relating events through causal evidence, CS and EU
summarize or organize information across a temporal segment, CT accumulates
repeated observations over the stream, and ACU relies on surrounding context
to identify anomalous situations. FlashBack improves causal reasoning and
counting on both backbones, while the event-understanding gain is concentrated
on Mage-VL. Clip summarization and anomaly-context understanding remain stable
or improve, indicating that historical access can support several forms of
temporal integration without uniformly benefiting every StreamingBench task.
Together with the mixed group-level Contextual results, these effects suggest
that the value of memory depends on whether prior observations provide
supporting evidence or instead introduce context that is no longer relevant to
the current query.

\par
\WFclear
\endgroup

Across these results, FlashBack improves representative
long-range temporal tasks while largely preserving Native real-time
performance. Complete results are provided in
Appendix~\ref{app:detailed-results}.
\leavevmode\par
\vspace{3pt}
\begin{figure}[h]
    \centering
    \includegraphics[width=1\linewidth]{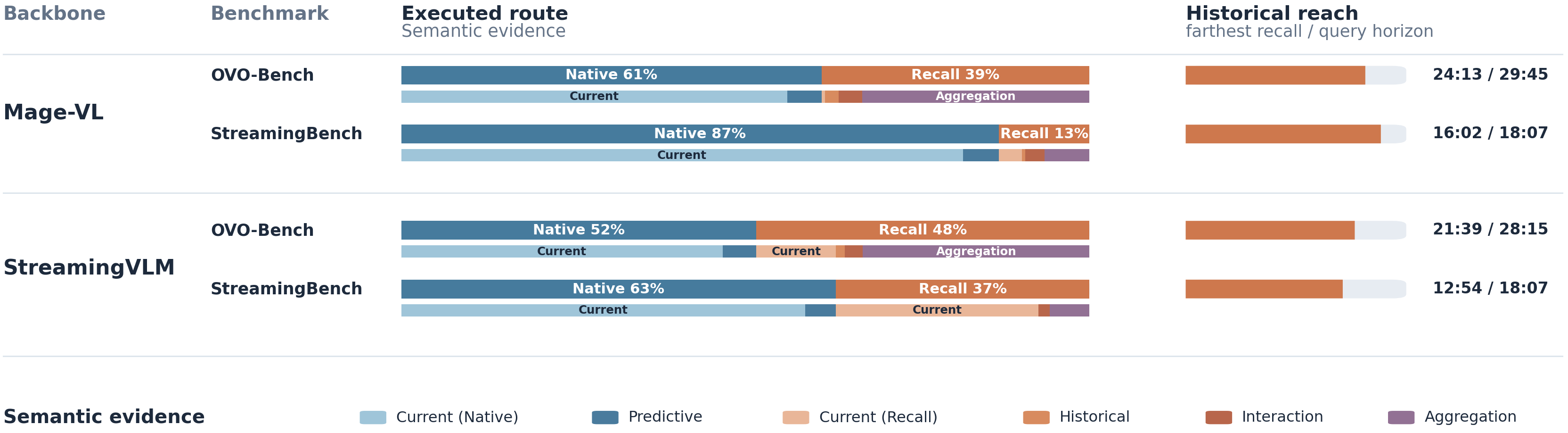}
    \vspace{-4pt}
    \caption{Routing behavior and historical reach of FlashBack.}
    \label{fig:routing-profile}
\end{figure}

\noindent\textbf{Routing behavior and historical reach.}
Figure~\ref{fig:routing-profile} summarizes the routing semantics and
historical reach of FlashBack. Current queries may map to either trajectory
after refinement, Predictive queries remain on the Native trajectory, and
Historical, Interaction, and Aggregation requirements invoke Recall. The historical-reach profiles further show that Recall is not confined to the
recent context: for both backbones, selected pages can originate more than
20 minutes before the query. Together, these views reflect the central design
of FlashBack: history participates only when required by the query, while
Recall provides access to long-range evidence beyond the bounded Native state.


\begingroup
\setlength{\intextsep}{0pt}
\setlength{\abovecaptionskip}{0pt}
\setlength{\belowcaptionskip}{2pt}
\setlength{\columnsep}{8pt}

\begin{wrapfigure}{r}{0.7\columnwidth}
\vspace{-8pt}
\centering

\includegraphics[
    width=\linewidth
]{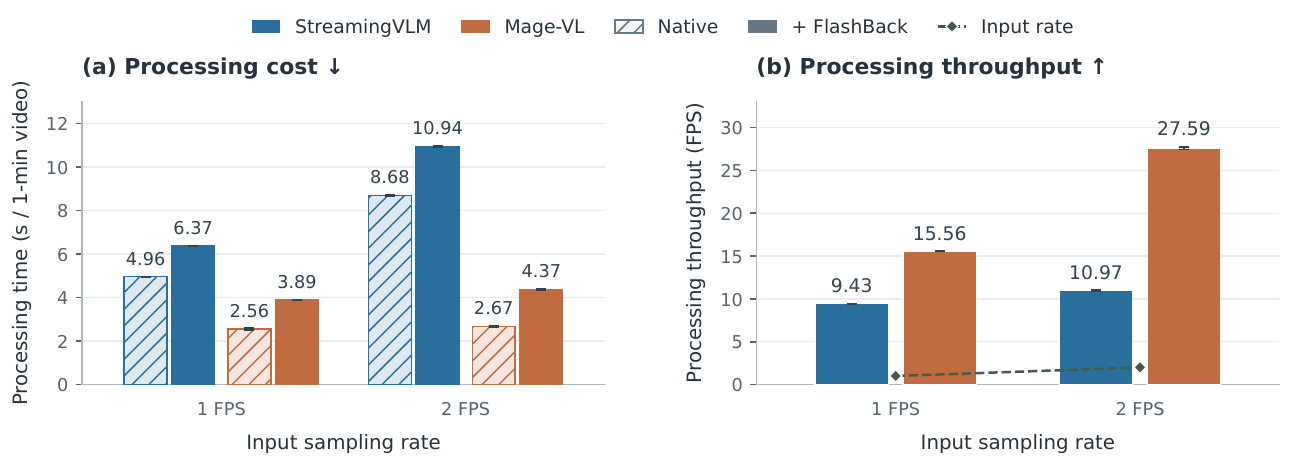}

\caption{Efficiency of FlashBack.}
\label{fig:efficiency}

\end{wrapfigure}

\noindent
\noindent\textbf{Efficiency Analysis.} Figure~\ref{fig:efficiency} shows the computational overhead of FlashBack.
This increases processing time relative to the Native trajectory, but both
backbones still process the stream faster than it arrives at 1 and 2 FPS.
Their throughput remains several times higher than the corresponding input
rate, leaving clear processing headroom under both evaluated settings.
Detailed measurement protocols are provided in
Appendix~\ref{app:efficiency}.

\par
\WFclear
\endgroup

\begin{figure}[h]
    \centering
    \includegraphics[width=1\linewidth]{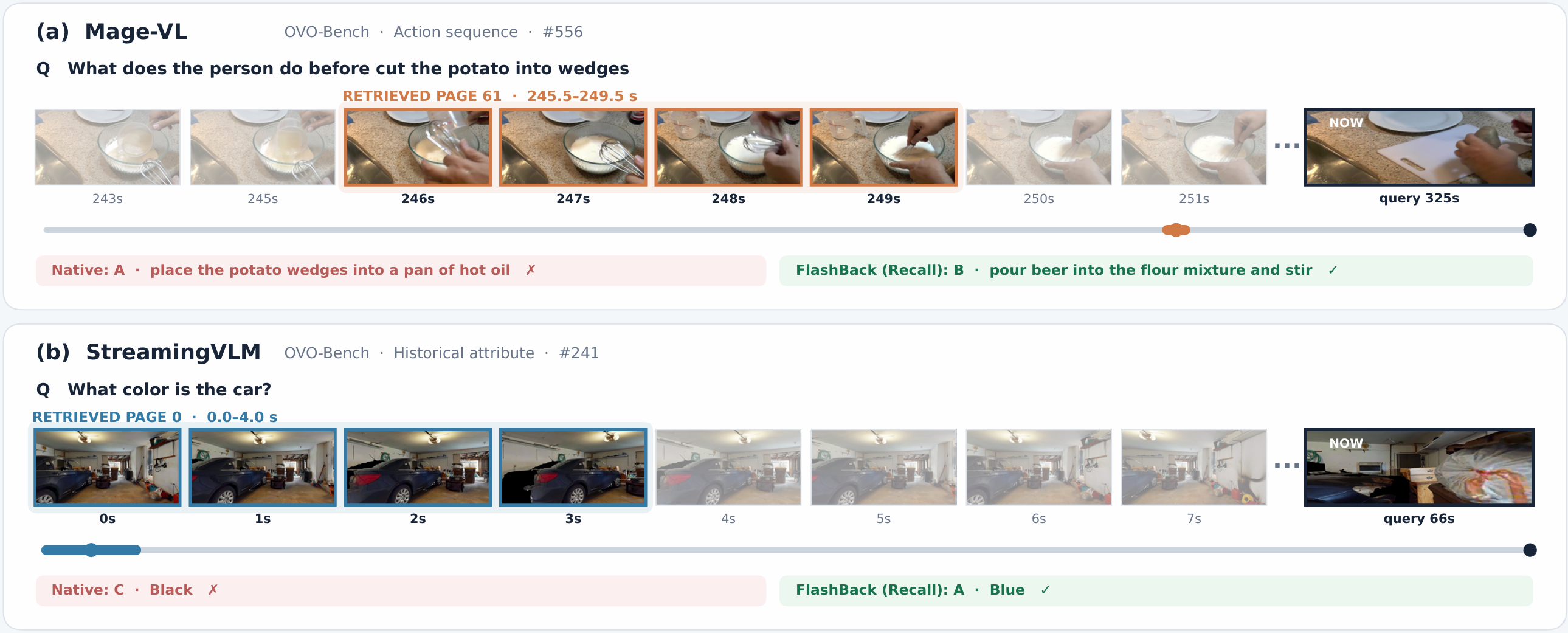}
    \vspace{-16pt}
    \caption{Representative FlashBack visualization cases on OVO-Bench.}
    \label{fig:qualitative}
\end{figure}

\noindent\textbf{Visualization Cases.} Figure~\ref{fig:qualitative} shows representative cases in which the Native
trajectory misses earlier evidence, whereas Recall retrieves the relevant
historical page and supports the correct answer on both backbones. Additional examples are provided in Appendix~\ref{app:qualitative}.
\vspace{1mm}

\section{Conclusion}
We introduced FlashBack, a training-free framework that selectively accesses
long-term history through temporal-evidence routing and an isolated Side-KV
Recall trajectory. Across two streaming backbones and two benchmarks, it
improves multiple tasks requiring historical recall and long-range temporal
evidence while largely preserving Native real-time capability, achieving
competitive performance without additional training.

\section{Limitations}
FlashBack is training-free, so neither the temporal-evidence router nor the
memory retriever is explicitly optimised for downstream streaming tasks. Its
benefits are query dependent: historical access supports queries
requiring long-range evidence, but can introduce interference in
hallucination-sensitive tasks that require rejecting unsupported premises and
context-filtering tasks that require distinguishing the current state from
stale history. Future work may address these cases through uncertainty- and
validity-aware routing and retrieval that suppress unsupported or temporally
outdated evidence while preserving the isolated Recall pathway and
frozen-backbone setting.

\bibliography{references}

@inproceedings{fu2025video,
  title={Video-mme: The first-ever comprehensive evaluation benchmark of multi-modal llms in video analysis},
  author={Fu, Chaoyou and Dai, Yuhan and Luo, Yongdong and Li, Lei and Ren, Shuhuai and Zhang, Renrui and Wang, Zihan and Zhou, Chenyu and Shen, Yunhang and Zhang, Mengdan and others},
  booktitle={2025 IEEE/CVF Conference on Computer Vision and Pattern Recognition (CVPR)},
  pages={24108--24118},
  year={2025},
  organization={IEEE}
}

@inproceedings{chen2024videollm,
  title={Videollm-online: Online video large language model for streaming video},
  author={Chen, Joya and Lv, Zhaoyang and Wu, Shiwei and Lin, Kevin Qinghong and Song, Chenan and Gao, Difei and Liu, Jia-Wei and Gao, Ziteng and Mao, Dongxing and Shou, Mike Zheng},
  booktitle={2024 IEEE/CVF Conference on Computer Vision and Pattern Recognition (CVPR)},
  pages={18407--18418},
  year={2024},
  organization={IEEE}
}

@inproceedings{niu2025ovo,
  title={Ovo-bench: How far is your video-llms from real-world online video understanding?},
  author={Niu, Junbo and Li, Yifei and Miao, Ziyang and Ge, Chunjiang and Zhou, Yuanhang and He, Qihao and Dong, Xiaoyi and Duan, Haodong and Ding, Shuangrui and Qian, Rui and others},
  booktitle={2025 IEEE/CVF Conference on Computer Vision and Pattern Recognition (CVPR)},
  pages={18902--18913},
  year={2025},
  organization={IEEE}
}

@inproceedings{xu2026streamingvlm,
  title={Streamingvlm: Real-time understanding for infinite video streams},
  author={Xu, Ruyi and Xiao, Guangxuan and Chen, Yukang and He, Liuning and Peng, Kelly and Lu, Yao and Han, Song},
  booktitle={International Conference on Learning Representations},
  volume={2026},
  pages={61463--61475},
  year={2026}
}

@article{shen2026simple,
  title={A simple baseline for streaming video understanding},
  author={Shen, Yujiao and Tian, Shulin and Yang, Jingkang and Liu, Ziwei},
  journal={arXiv preprint arXiv:2604.02317},
  year={2026}
}

@inproceedings{zhang2025flash,
  title={Flash-vstream: Efficient real-time understanding for long video streams},
  author={Zhang, Haoji and Wang, Yiqin and Tang, Yansong and Liu, Yong and Feng, Jiashi and Jin, Xiaojie},
  booktitle={2025 IEEE/CVF International Conference on Computer Vision (ICCV)},
  pages={21059--21069},
  year={2025},
  organization={IEEE}
}

@inproceedings{di2025streaming,
  title={Streaming video question-answering with in-context video kv-cache retrieval},
  author={Di, Shangzhe and Yu, Zhelun and Zhang, Guanghao and Li, Haoyuan and Zhong, Tao and Cheng, Hao and Li, Bolin and He, Wanggui and Shu, Fangxun and Jiang, Hao},
  booktitle={International Conference on Learning Representations},
  volume={2025},
  pages={42115--42127},
  year={2025}
}

@inproceedings{zhang2026hermes,
  title={Hermes: Kv cache as hierarchical memory for efficient streaming video understanding},
  author={Zhang, Haowei and Yang, Shudong and Fu, Jinlan and Ng, See Kiong and Qiu, Xipeng},
  booktitle={Proceedings of the 64th Annual Meeting of the Association for Computational Linguistics (Volume 1: Long Papers)},
  pages={8411--8430},
  year={2026}
}

@misc{ning2025livevlm,
      title={LiveVLM: Efficient Online Video Understanding via Streaming-Oriented KV Cache and Retrieval}, 
      author={Zhenyu Ning and Guangda Liu and Qihao Jin and Chengwei Li and Wenchao Ding and Minyi Guo and Jieru Zhao},
      year={2026},
      eprint={2505.15269},
      archivePrefix={arXiv},
      primaryClass={cs.CV},
      doi={https://doi.org/10.1145/3770743.3804012},
}

@InProceedings{liang2026oasis,
    author    = {Liang, Zhijia and Li, Jiaming and Chen, Weikai and Zhang, Yanhao and Lu, Haonan and Li, Guanbin},
    title     = {OASIS: On-Demand Hierarchical Event Memory for Streaming Video Reasoning},
    booktitle = {Proceedings of the IEEE/CVF Conference on Computer Vision and Pattern Recognition (CVPR)},
    month     = {June},
    year      = {2026},
    pages     = {2821-2831}
}

@article{wu2026semantic,
  title={Semantic-Aware Adaptive Visual Memory for Streaming Video Understanding},
  author={Wu, Hang and Mathews, Sherin Mary and Cai, Yujun and Yang, Ming-Hsuan and Wang, Yiwei},
  journal={arXiv preprint arXiv:2605.07897},
  year={2026}
}

@article{ge2026should,
  title={What Should a Streaming Video Model Remember?},
  author={Ge, Haonan and Wang, Yiwei and Wu, Hang and Cai, Yujun},
  journal={arXiv preprint arXiv:2606.16353},
  year={2026}
}

@article{fan2026folio,
  title={FOLIO: Focused Semantic Memory for Streaming Video Understanding},
  author={Fan, Haoyang and Parikh, Dhruv and Ramachandran, Anvitha and Gobriel, Sameh and Jain, Nilesh and Kannan, Rajgopal and Prasanna, Viktor},
  journal={arXiv preprint arXiv:2607.13298},
  year={2026}
}

@inproceedings{long2026seeing,
  title={Seeing, listening, remembering, and reasoning: A multimodal agent with long-term memory},
  author={Long, Lin and He, Yichen and Ye, Wentao and Pan, Yiyuan and Lin, Yuan and Li, Hang and Zhao, Junbo and Li, Wei},
  booktitle={International Conference on Learning Representations},
  volume={2026},
  pages={146197--146246},
  year={2026}
}

@article{wen2026eventmemagent,
  title={Eventmemagent: Hierarchical event-centric memory for online video understanding with adaptive tool use},
  author={Wen, Siwei and Wang, Zhangcheng and Zhang, Xingjian and Huang, Lei and Wu, Wenjun},
  journal={arXiv preprint arXiv:2602.15329},
  year={2026}
}

@article{li2026visual,
  title={Visual Agentic Memory: Enabling Online Long Video Understanding via Online Indexing, Hierarchical Memory, and Agentic Retrieval},
  author={Li, Aiden Yiliu and Numan, Nels and Steed, Anthony},
  journal={arXiv preprint arXiv:2605.16481},
  year={2026}
}

@inproceedings{huang2025online,
  title={Online video understanding: Ovbench and videochat-online},
  author={Huang, Zhenpeng and Li, Xinhao and Li, Jiaqi and Wang, Jing and Zeng, Xiangyu and Liang, Cheng and Wu, Tao and Chen, Xi and Li, Liang and Wang, Limin},
  booktitle={2025 IEEE/CVF Conference on Computer Vision and Pattern Recognition (CVPR)},
  pages={3328--3338},
  year={2025},
  organization={IEEE}
}

@article{yang2026mage,
  title={Mage-VL: An Efficient Codec-Native Streaming Multimodal Foundation Model},
  author={Yang, Senqiao and Zhang, Kaichen and Jia, Zhaoyang and Guo, Jinghao and Shen, Yifei and Zhang, Xinjie and Zhang, Xiaoyi and Wang, Haoqing and Li, Xiao and Zhang, Peng and others},
  journal={arXiv preprint arXiv:2607.24904},
  year={2026}
}

@inproceedings{xie2026streamrag,
  title={StreamRAG: Enhancing Real-Time Video Understanding with Retrieval Augmentation},
  author={Xie, Junlin and Zheng, Quanlong and Zhang, Ruifei and Wang, Kuo and Zhang, Yanhao and Luo, Jinguo and Lu, Haonan and Wan, Xiang and Li, Guanbin},
  booktitle={Proceedings of the IEEE/CVF Conference on Computer Vision and Pattern Recognition},
  pages={38870--38879},
  year={2026}
}

@inproceedings{dai2019transformer,
  title={Transformer-xl: Attentive language models beyond a fixed-length context},
  author={Dai, Zihang and Yang, Zhilin and Yang, Yiming and Carbonell, Jaime G and Le, Quoc and Salakhutdinov, Ruslan},
  booktitle={Proceedings of the 57th annual meeting of the association for computational linguistics},
  pages={2978--2988},
  year={2019}
}

@inproceedings{xiao2024efficient,
  title={Efficient streaming language models with attention sinks},
  author={Xiao, Guangxuan and Tian, Yuandong and Chen, Beidi and Han, Song and Lewis, Mike},
  booktitle={International Conference on Learning Representations},
  volume={2024},
  pages={21875--21895},
  year={2024}
}

@inproceedings{
kim2025infinipotv,
title={InfiniPot-V: Memory-Constrained {KV} Cache Compression for Streaming Video Understanding},
author={Minsoo Kim and Kyuhong Shim and Jungwook Choi and Simyung Chang},
booktitle={The Thirty-ninth Annual Conference on Neural Information Processing Systems},
year={2025},
}

@inproceedings{lin2026streamingbench,
  title={Streamingbench: Assessing the gap for mllms to achieve streaming video understanding},
  author={Lin, Junming and Fang, Zheng and Chen, Chi and Cheng, Haoxuan and Wan, Zihao and Luo, Fuwen and Wang, Ziyue and Li, Peng and Liu, Yang and Sun, Maosong},
  booktitle={ICASSP 2026-2026 IEEE International Conference on Acoustics, Speech and Signal Processing (ICASSP)},
  pages={12147--12151},
  year={2026},
  organization={IEEE}
}

@inproceedings{azad2026streamready,
  title={Streamready: Learning what to answer and when in long streaming videos},
  author={Azad, Shehreen and Vineet, Vibhav and Rawat, Yogesh S},
  booktitle={Proceedings of the IEEE/CVF Conference on Computer Vision and Pattern Recognition},
  pages={40494--40504},
  year={2026}
}

@article{hao2026shallowstream,
  title={ShallowStream: Index Shallow then Answer Deep for Streaming Video Understanding},
  author={Hao, Jitai and Yang, Ke and Huang, Qiang and Yu, Jun},
  journal={arXiv preprint arXiv:2609.02780},
  year={2026}
}
\bibliographystyle{unsrtnat}

\appendix
\section{Additional Implementation Details}
\label{app:implementation}

\subsection{Component Ablation Settings}

All component ablations are conducted on StreamingVLM with the released checkpoint and frozen model parameters. We use 4-second streaming units and maintain a 16-second recent context. Decoder-level candidate reranking uses layers 10--17 under zero-based
indexing, while the Recall trajectory is executed through all 28 decoder
layers. Unless otherwise specified, reranking selects $K=1$ long-term unit. For the memory-hierarchy ablation in Table~\ref{tab:ablation-memory},
Structured Routing is kept fixed while varying \(K\).

The ablation experiments use the same video sampling, query inputs, answer
prefix, answer-candidate space, and scoring rule across all compared variants. Each ablation
then varies the corresponding memory or routing component under study. The two diagnostic subsets are StreamingBench Multimodal Alignment and OVO-Bench Clues Reveal Responding, as described in Section~\ref{sec:datasets}.

For finite-answer evaluation, we use next-token candidate logits after a fixed answer prefix rather than unrestricted generation. Native and all ablation variants share the same candidate coordinates and tie-breaking rule for each question.

\subsection{Main Experiment Settings}
\label{app:main setting}
\paragraph{StreamingVLM.}
StreamingVLM uses the released checkpoint and its original streaming-state management. We use 4-second memory units and a 16-second recent context. Decoder-level candidate reranking uses layers 10--17 under zero-based
indexing, and Recall is processed through the complete 28-layer decoder. Video input is sampled at 2 FPS on OVO-Bench and StreamingBench.

\paragraph{Mage-VL-4B.}
Mage-VL-4B uses the released model with the frame-input streaming interface. FlashBack organizes memory using 4-frame units and a 16-frame recent context. Decoder-level candidate reranking uses layers 14--21 under zero-based
indexing, while Recall is processed through all 36 decoder layers. Mage-VL-4B uses 1 FPS input on OVO-Bench and StreamingBench.

Within every backbone--benchmark pair, Native and FlashBack use identical video sampling, query inputs, and answer-scoring settings. The different sampling rates across StreamingVLM and Mage-VL-4B reflect their corresponding evaluation adapters and are not treated as a controlled FPS comparison.

\subsection{Answer Scoring}
\label{app:scoring}
We use a deterministic candidate-logit protocol for finite-answer tasks. Given a fixed answer prefix, we obtain the next-token logits of the permitted answer candidates and select the candidate with the highest value. This protocol reduces variation arising from free-form instruction following and answer formatting and follows the evaluation interface used with StreamingVLM.

The candidate spaces are benchmark-specific. OVO-Bench uses a fixed 17-token space
\[
\{\mathrm{No},\mathrm{Yes},0,\ldots,9,\mathrm{A},\ldots,\mathrm{E}\},
\]
whereas StreamingBench uses the option labels presented by each multiple-choice question. Native and FlashBack always share the same prompt, answer prefix, candidate coordinates, and tie-breaking rule for a given question.

All reported experiments use visual input only; no additional audio branch is introduced.

\subsection{Temporal-Evidence Routing}
\label{app:router}

FlashBack selects between the Native and Recall trajectories at query time
according to the temporal evidence required by the request. The temporal-evidence router uses the textual reasoning capability of the frozen
backbone and introduces no additional training. The temporal-evidence states are
inference-time semantic variables; no external routing annotations are used. Its objective is to identify
the temporal support requested by the query rather than to predict which
trajectory is more likely to produce a correct answer. Both backbones use the
same semantic definitions, routing prompts, and fixed semantic-to-action
mappings.

\paragraph{Router input.}
Both routing strategies receive the same normalized question-only input derived
from the current query \(q_t\). Temporal qualifiers and any outer
evidence-sufficiency structure are retained, while instructions that only
constrain answer formatting are removed. Neither router receives video frames,
stored visual memory, answer candidates, branch predictions, candidate logits,
retrieval scores, or answer labels.

\paragraph{Structured temporal-evidence routing.}
The first stage assigns the request to one of the five temporal-evidence states
defined in the main text: \emph{Current}, \emph{Historical},
\emph{Interaction}, \emph{Aggregation}, or \emph{Predictive}. The
prompt-facing labels
\pcode{CURRENT\_PERCEPTION},
\pcode{HISTORICAL\_STATE},
\pcode{TEMPORAL\_INTERACTION},
\pcode{GLOBAL\_AGGREGATION}, and
\pcode{PREDICTIVE\_CONSEQUENCE}
correspond to these five states, respectively. The boxes below show the
canonical prompt views. Typographic emphasis is added only for readability;
the underlying prompt wording is unchanged.

\begin{promptbox}{Stage-1 temporal-evidence prompt}

Classify the \pkey{temporal proof form} requested by the visual question.
Classify only the \pkey{request semantics}, without assuming facts that are
not stated in the request.

\medskip
\pkey{Apply this priority:}
prefix-wide quantification or coverage is
\pcode{GLOBAL\_AGGREGATION};
an explicitly unobserved future target is
\pcode{PREDICTIVE\_CONSEQUENCE};
an observed order, change, interaction, or cause is
\pcode{TEMPORAL\_INTERACTION};
one explicitly earlier bounded fact is
\pcode{HISTORICAL\_STATE};
otherwise a query-boundary or latest-current target is
\pcode{CURRENT\_PERCEPTION}.

\medskip
A word such as \emph{count}, \emph{compare}, \emph{before}, \emph{after},
\emph{first}, \emph{last}, or \emph{never} has no class by itself; use the
time domain over which the question applies it.

\medskip
\pkey{Contrast the following proof forms:}
a property requested at an explicit earlier time is
\pcode{HISTORICAL\_STATE};
the same property requested at the query boundary is
\pcode{CURRENT\_PERCEPTION};
an observed transition is
\pcode{TEMPORAL\_INTERACTION};
a statistic over the observed prefix is
\pcode{GLOBAL\_AGGREGATION};
and an unobserved later target is
\pcode{PREDICTIVE\_CONSEQUENCE}.

\medskip
\pkey{Choose exactly one routing code.}

\medskip
\textbf{A.} \pcode{CURRENT\_PERCEPTION}: 
The request targets a configuration, state, or activity at the query boundary,
including a latest state treated as current. It is not an explicitly earlier
fact, an event relation, a prefix-wide aggregate, or an unobserved future
target.

\smallskip
\textbf{B.} \pcode{HISTORICAL\_STATE}: 
The request targets one explicitly earlier, completed, bounded event or state.
It does not ask for a relation among events or for coverage over multiple
moments.

\smallskip
\textbf{C.} \pcode{TEMPORAL\_INTERACTION}: 
The request targets an observed order, transition, change, interaction, or
causal relation inside one bounded episode.

\smallskip
\textbf{D.} \pcode{GLOBAL\_AGGREGATION}: 
The request targets multiple moments or the observed prefix as a set:
aggregation, repetition, frequency, cross-period comparison, coverage,
exclusion, first/last/unique, ever/never, or absence over time.

\smallskip
\textbf{E.} \pcode{PREDICTIVE\_CONSEQUENCE}: 
The requested outcome lies after the query boundary and has not yet been
observed; the request asks for a forecast rather than an already realized
result.

\medskip
\pkey{Visual request:}

\pcode{<ROUTER\_INPUT>}

\smallskip
\pkey{Question type:}

\end{promptbox}

Only requests assigned to \emph{Current} enter the second semantic stage.
This stage distinguishes an immediately observable boundary configuration,
an ongoing process, a state established by earlier evidence and persisting to
the current time, and the realized outcome of a bounded process. These
correspond to the formal states \emph{Boundary}, \emph{Ongoing},
\emph{Persistent}, and \emph{Outcome}, respectively. The same normalized
request is used in both stages.

\begin{promptbox}{Stage-2 Current refinement prompt}

Classify the proof form of this
\pcode{CURRENT\_PERCEPTION} request.
Classify only the \pkey{target semantics}, without assuming facts that are not
stated in the request.

\medskip
\pkey{Explicit wording controls the distinction:}
currently performing or moving is
\pcode{ONGOING\_PROCESS};
explicitly visible, present, or configured at this moment is
\pcode{SIMULTANEOUS\_BOUNDARY\_CONFIGURATION};
latest-established, latest-known, or based on a prior update is
\pcode{PERSISTENT\_CURRENT\_STATE}.

\medskip
\pkey{Apply this priority:}
an already-realized result is
\pcode{REALIZED\_BOUNDED\_OUTCOME};
an activity still underway is
\pcode{ONGOING\_PROCESS};
a state that semantically persists until updated is
\pcode{PERSISTENT\_CURRENT\_STATE};
otherwise an instantaneous configuration is
\pcode{SIMULTANEOUS\_BOUNDARY\_CONFIGURATION}.

\medskip
\pkey{Choose exactly one routing code.}

\medskip
\textbf{S.} \pcode{SIMULTANEOUS\_BOUNDARY\_CONFIGURATION}: 
The request explicitly targets what is visually present, visible, or configured
right at the query boundary: appearance, visible text, a local quantity, or a
simultaneous spatial relation. Use this for an explicit current observation
even if that observed property could physically persist.

\smallskip
\textbf{D.} \pcode{ONGOING\_PROCESS}: 
The target is an action, motion, gesture, or process still underway at the
query boundary; it does not ask for its completed result.

\smallskip
\textbf{L.} \pcode{PERSISTENT\_CURRENT\_STATE}: 
The request explicitly asks for the latest established or carried state
determined by a prior update, handoff, or observation, such as latest-known
location or possession. Do not choose this merely because a property visible
at the boundary could persist.

\smallskip
\textbf{T.} \pcode{REALIZED\_BOUNDED\_OUTCOME}: 
The target is the already-realized completion, continuation result, or outcome
of a bounded process that began earlier; it is neither ongoing nor an
unobserved future forecast.

\medskip
\pkey{Visual request:}

\pcode{<ROUTER\_INPUT>}

\smallskip
\pkey{CURRENT proof form:}

\end{promptbox}

The resulting semantic state is converted to an inference action using the
fixed mappings described in the main text. At the first stage,
\emph{Historical}, \emph{Interaction}, and \emph{Aggregation} invoke Recall,
whereas \emph{Predictive} remains on the Native trajectory.
\emph{Current} requests enter the second stage, where \emph{Boundary} and
\emph{Ongoing} map to Native, while \emph{Persistent} and \emph{Outcome} map
to Recall. Recall is an inference action rather than a statement that eligible
long-term history must be available. When $\mathcal{E}_t=\varnothing$, the
Recall trajectory uses the causally available recent context $L_t$, as defined
in the main text.

\paragraph{Representation-balanced inference.}
Although the semantic definitions remain fixed, a single categorical prompt
can be sensitive to the position of a definition and to its arbitrary output
code. We therefore instantiate representation-balanced inference using cyclic
prompt views. The first stage uses five views and the second uses four, such
that every semantic state appears exactly once at each definition position and
with each output code.

For stage $k$, let $V^{(k)}$ denote its set of balanced views,
$P_v^{(k)}(q_t)$ the prompt instantiated under view $v$, and
$z_v^{(k)}(c)$ the output code assigned to semantic state $c$ in that view.
Let $\boldsymbol{\ell}_v^{(k)}(q_t)$ denote the next-token logits over the
candidate output codes. For each view, we apply a log-softmax over these
candidate logits and map the resulting scores back to the shared semantic
space. The aggregated score is

\[
\bar{s}^{(k)}(c \mid q_t)
=
\frac{1}{|V^{(k)}|}
\sum_{v \in V^{(k)}}
\log \operatorname{softmax}
\!\left(
\boldsymbol{\ell}_v^{(k)}(q_t)
\right)_{z_v^{(k)}(c)},
\qquad
\hat{c}_t^{(k)}
=
\arg\max_{c \in \mathcal{C}^{(k)}}
\bar{s}^{(k)}(c \mid q_t).
\]

We use $|V^{(1)}|=5$ and $|V^{(2)}|=4$. The aggregation therefore averages
the mapped candidate-code log probabilities rather than performing majority
voting or taking an arithmetic mean of the raw probabilities. No
benchmark-specific class weights or decision thresholds are used.

\paragraph{Direct routing baseline.}
The Direct Routing baseline in Table~1 receives the same normalized request
but removes the intermediate temporal-evidence semantics and directly predicts
either Native or Recall. Its canonical prompt view is shown below.

\begin{promptbox}{Direct Native--Recall routing prompt}

Choose how a streaming video assistant should obtain evidence for the request.
Choose directly between the two paths using the
\pkey{evidence requirement expressed in the request}.

\medskip
\pkey{Do not answer the question or predict which path will produce a correct
answer.}
The decision has no access to the video or the contents of stored memory.
Judge the complete request, not a single keyword.

\medskip
\pkey{Output exactly one code.}

\medskip
\textbf{A.} \pcode{NATIVE}: 
Answer through the original live streaming state, without a separate historical
lookup, for a question grounded in the immediate scene or ongoing process,
including a forecast from the currently unfolding situation.

\smallskip
\textbf{B.} \pcode{RECALL}: 
Consult stored visual history together with recent observations, for a question
requiring a past observation, a carried-over state or completed outcome,
relationships across time, or accumulated evidence over the observed stream.

\medskip
\pkey{Request:}

\pcode{<ROUTER\_INPUT>}

\smallskip
\pkey{Path:}

\end{promptbox}

Direct Routing uses two balanced views that exchange both the category order
and the category-to-code assignment. Their candidate-code log probabilities
are mapped back to the shared Native/Recall space and averaged in the same
manner as in the structured router. This baseline tests whether the routing
decision can be made directly in the Native/Recall action space without the
intermediate temporal-evidence decomposition used by FlashBack.

For the routing-policy ablation, the Native and Recall branch outputs are held
fixed for each query, and only the textual routing decision is recomputed.
Each routing policy then selects between the same two branch outputs. This
isolates the effect of the routing policy from differences in branch
computation and answer scoring.

\subsection{Trajectory-Selection Analysis}
\label{app:trajectory-selection}

The temporal-evidence states used by FlashBack are inference-time semantic
variables rather than supervised task labels. We therefore evaluate the router
through its ability to select a trajectory that produces the correct downstream
answer, rather than through agreement with manually defined routing labels.

\paragraph{Evaluation protocol.}
For every query, we compute one Native output and one Recall output and retain
both throughout the analysis. Recall uses the complete FlashBack pipeline,
including causal history, candidate prefiltering, decoder-level Q--K reranking
with \(K=1\), and Side-KV. Native, Always Recall, Direct Routing, Structured
Routing, and the Trajectory Oracle are all evaluated using these same fixed
trajectory outputs, without rerunning video inference.

Direct and Structured Routing receive the same normalized question-only input.
Their decisions are produced independently of the trajectory predictions,
candidate logits, retrieval scores, correctness outcomes, and answer labels.
When no eligible historical page exists, Recall uses the recent context \(L_t\)
through the standard recent-context fallback. For short observed prefixes, \(L_t\)
may cover the complete prefix, causing the fixed Recall output to coincide
numerically with the Native output for some queries. Correctness is determined
using the same finite-candidate next-token logit scoring as in the main
experiments, without an additional judge or routing-specific correctness
criterion.
\vspace{4pt}
\begin{table*}[h]
\centering
\captionsetup{
    font=footnotesize,
    labelfont=normalfont,
    textfont=bf,
    skip=3pt
}
\caption{\textbf{Outcome-based trajectory-selection analysis.}
The correctness partition is derived from the same fixed Native and Recall
outputs used by all selection policies.}
\label{tab:trajectory-selection}
\vspace{-1mm}

\setlength{\tabcolsep}{8pt}
\renewcommand{\arraystretch}{1.06}

\resizebox{0.72\linewidth}{!}{%
\begin{tabular}{@{}lcc@{}}
\toprule
\textbf{Selection policy or diagnostic}
& \textbf{MA}
& \textbf{CRR} \\
\midrule

\multicolumn{3}{@{}l}{\color{traintext}\textit{Downstream accuracy (\%)}} \\

Native
& 69.60 & 41.67 \\

Always Recall
& 70.40 & 60.83 \\

Direct Routing
& 69.20 & 41.67 \\

Structured Routing
& \textbf{72.80} & \textbf{60.83} \\

Trajectory Oracle
& 78.80 & 65.42 \\

\midrule
\multicolumn{3}{@{}l}{\color{traintext}\textit{Correctness partition (number of queries)}} \\

Native only
& 21 & 11 \\

Recall only
& 23 & 57 \\

Both
& 153 & 89 \\

Neither
& 53 & 83 \\

\midrule
\multicolumn{3}{@{}l}{\color{traintext}\textit{Trajectory-selection performance}} \\

Direct: unique-correct selection
& 20/44 (45.45\%) & 11/68 (16.18\%) \\

Structured: unique-correct selection
& \textbf{29/44 (65.91\%)}
& \textbf{57/68 (83.82\%)} \\
\midrule
Direct: answerable-query success
& 173/197 (87.82\%)
& 100/157 (63.69\%) \\

Structured: answerable-query success
& \textbf{182/197 (92.39\%)}
& \textbf{146/157 (92.99\%)} \\

\bottomrule
\end{tabular}%
}

\vspace{-2mm}
\end{table*}

\paragraph{Outcome-derived correctness partition.}
Each query belongs to one of four partitions according to the correctness of
its two fixed trajectory outputs: \emph{Native only}, \emph{Recall only},
\emph{Both}, or \emph{Neither}. This partition is derived from downstream
answer correctness rather than from manually annotated routing labels. It
characterizes whether trajectory selection can affect the final answer:
selection is consequential for Native-only and Recall-only queries, either
trajectory succeeds for Both queries, and neither fixed trajectory succeeds
for Neither queries.

The Trajectory Oracle is correct whenever at least one of the two fixed
trajectory outputs is correct,
\[
\operatorname{Oracle}(q_t)
=
\mathbb{I}
\left[
y_t^{N}=y_t
\;\lor\;
y_t^{R}=y_t
\right],
\]
where \(y_t^{N}\) and \(y_t^{R}\) are the Native and Recall predictions and
\(y_t\) is the benchmark answer. It therefore provides an analytical upper
bound for selection between the two available outputs and is not used during
inference. Structured Routing is \(6.00\) percentage points below this bound
on MA and \(4.58\) points below it on CRR.

\paragraph{Trajectory-selection performance.}
For queries where exactly one trajectory is correct, we measure whether the
router selects that unique correct trajectory. Structured Routing substantially
improves this selection rate over Direct Routing on both subsets, with the
largest difference on CRR.

We additionally evaluate selection success over queries answerable by at least
one fixed trajectory:
\begin{equation}
\mathrm{SelectionSuccess}
=
\frac{
N_{\mathrm{both}}
+
N_{\mathrm{unique\ selected}}
}{
N_{\mathrm{native\ only}}
+
N_{\mathrm{recall\ only}}
+
N_{\mathrm{both}}
}.
\end{equation}
Both-correct queries are counted as successful because either trajectory
produces the correct answer, while Neither queries are excluded because no
selection policy can recover a correct answer from the two fixed outputs.
Structured Routing selects a correct trajectory for more than \(92\%\) of
these answerable queries on both MA and CRR. Together with the downstream
accuracy results, this analysis shows that the structured temporal-evidence
decomposition improves the practical Native/Recall decision without requiring
external routing annotations.

\subsection{Long-Term Retrieval Implementation}
\label{app:retrieval}
\paragraph{Causal pages and candidate construction.}
In the implementation, each historical unit
\(h_i=(R_i,\Delta_i)\) corresponds to one completed visual page together
with its temporal interval, rather than to an individual visual token.
StreamingVLM forms a page from a video chunk of up to four seconds, whereas
Mage-VL forms a page from up to four sampled frames. As defined in the main
text, only completed pages preceding the recent context belong to the eligible
set \(\mathcal E_t\).

Before decoder-level reranking, both backbones use a frozen,
question-conditioned prefilter to construct a small candidate set. Let
\(\mathcal X_t\) contain the question itself and, when answer options are
available, the corresponding question--option hypotheses. For a hypothesis
\(x\in\mathcal X_t\), let \(e_u\) denote the frozen model input embedding of
its \(u\)-th text token, and let \(V_i\) denote the compact visual states used
to represent page \(i\). We compute the token-level late-interaction score
\begin{equation}
a_\theta(x,h_i)
=
\frac{1}{|x|}
\sum_{u\in x}
\max_{v\in V_i}
\left\langle
\operatorname{norm}(e_u),
\operatorname{norm}(v)
\right\rangle,
\end{equation}
where
\(\operatorname{norm}(\mathbf{x})=\mathbf{x}/\|\mathbf{x}\|_2\).
The candidate set is then
\begin{equation}
\mathcal C_t
=
\begin{cases}
\displaystyle
\bigcup_{x\in\mathcal X_t}
\operatorname{Top1}_{h_i\in\mathcal E_t}
a_\theta(x,h_i),
&
\mathcal E_t\neq\varnothing,\\[6pt]
\varnothing,
&
\mathcal E_t=\varnothing.
\end{cases}
\end{equation}
Duplicate pages produced by different hypotheses are removed by the union.
Mage-VL represents each page in this prefilter using its initial, mean, and
final visual states, while StreamingVLM additionally includes a nonzero
temporal-difference state. All available answer options are treated
symmetrically; neither the ground-truth answer label nor the correctness of
either answering trajectory is used during candidate construction.

\paragraph{Backbone-internal page relevance.}
Candidates in \(\mathcal C_t\) are subsequently reranked in the frozen
decoder's query--key space. At retrieval layer \(\ell\), page \(i\) provides
visual key vectors
\(\{\mathbf{k}_{i,\ell,h,v}\}_{v=1}^{N_i}\)
for attention head \(h\). The corresponding query vector is taken from the
last answer-decision token of a memory-disabled query pass. After applying the
candidate-specific rotary coordinates, FlashBack scores a page by
\begin{equation}
s_\theta(q_t^R,L_t,h_i)
=
\frac{1}{|\mathcal L_\theta|}
\sum_{\ell\in\mathcal L_\theta}
\frac{1}{H_\ell}
\sum_{h=1}^{H_\ell}
\left[
\log
\sum_{v=1}^{N_i}
\exp\!\left(
\alpha_\ell
\bigl(\widetilde{\mathbf q}^{(i)}_{t,\ell,h}\bigr)^\top
\mathbf k_{i,\ell,h,v}
\right)
-
\log N_i
\right],
\end{equation}
where \(H_\ell\) is the number of attention heads at layer \(\ell\),
\(\alpha_\ell\) is the backbone attention scaling factor, and
\(\mathcal L_\theta\) is the set of retrieval layers. The score is therefore
a length-normalized log-sum-exp over the visual keys of each page, followed by
uniform averaging across attention heads and retrieval layers, rather than a
cosine similarity between pooled decoder states.

Using zero-based layer indices, we use layers \(10\)--\(17\) for
StreamingVLM and \(14\)--\(21\) for Mage-VL. For Mage-VL, the
memory-disabled query pass processes the current recent context \(L_t\),
followed by the question and answer prefix. For StreamingVLM, the query vector
is captured from the memory-disabled live-stream answer pass, while the current
window determines the canonical multimodal rotary coordinates of the
\textsc{Page}--\textsc{Current}--\textsc{Query} layout used when scoring
each candidate page.

The historical keys follow the native representation of each backbone.
StreamingVLM retains retrieval-layer KV states and the visual representations
required for later page construction as the stream progresses, while full-layer
page KV is materialized on demand for selected pages. Mage-VL
instead retains visual page fragments and materializes the exact decoder keys
for shortlisted pages when needed; these keys may subsequently be cached for
later queries. Neither implementation introduces a trained retrieval head.

\paragraph{Selection and empty history.}
The decoder-level scores are used in the main-text retrieval rule
\[
\mathcal R_t^K
=
\operatorname{TopK}_{h_i\in\mathcal C_t}
s_\theta(q_t^R,L_t,h_i).
\]
The main experiments use \(K=1\), with exact score ties resolved by the earlier
page identifier. For \(K>1\), if fewer than \(K\) candidates remain after
prefiltering, all available candidates are selected without padding or
expanding the search back to \(\mathcal E_t\). When
\(\mathcal E_t=\varnothing\), no long-term unit is attached and the Recall
trajectory uses the causally available recent context \(L_t\), as defined in
the main text. In the retained Mage-VL implementation, decoder-level Q--K
scoring is omitted when \(|\mathcal C_t|=1\), since reranking cannot change the
selected page.

\subsection{Side-KV Implementation and Hard-Insertion Controls}
\label{app:sidekv}

\paragraph{Query-local Side-KV construction.}
For Recall, FlashBack constructs an isolated decoder trajectory from the
retrieved historical page, the recent context, and the query, while preserving
the persistent Native trajectory. For StreamingVLM, the selected page provides
a stored or lazily materialized full-layer PageKV. The recent visual context is
prefetched independently from visual representations retained during online
processing. The historical-page and recent-context caches are then combined
into a query-local cache, allowing the query to attend jointly to both sources.
The two caches are constructed independently, so neither the historical page
nor the recent context is recomputed conditioned on the other. Multimodal RoPE
coordinates follow the canonical page--recent--query serialization, while the
historical page retains its original PageKV coordinates.

For Mage-VL, the selected historical page is materialized from its archived
representations into a separate query-local cache, followed by the recent
context and query at consecutive positions. In both backbones, the resulting
Recall trajectory is used only for answering the current query.

\paragraph{No Write-back and state lifetime.}
The retrieved page remains stored in long-term memory, while the decoder state
constructed for a Recall query is local to that query. KV states generated
while processing the recent context, query, and, when applicable, answer tokens
remain within the temporary Side-KV state and are released after the query is
completed. Subsequent streaming inference continues from the persistent Native
state. In the multiple-choice evaluations, answers are selected directly from
next-token candidate logits, so no answer-token KV is instantiated.

\paragraph{Hard-insertion controls.}
The hard-insertion controls in Table~\ref{tab:ablation-sidekv} are evaluated on
StreamingVLM with routing disabled. For every query with an eligible historical
page, all integration variants use the same causal candidate prefilter,
decoder-level Q--K reranking, and Top-1 historical-page selection. The complete
content span of the selected PageKV is used, while the shared chat header is
excluded because it is already present in the Native cache.

Each hard-insertion control appends the full-page tensors to a query-specific
clone of the persistent Native cache immediately before the answer forward.
This preserves the Native-cache insertion interface while keeping the evaluated
variants independent. \emph{Historical KV} directly reuses the selected page's
stored keys and values together with their canonical PageKV positions, without
re-reading the video or recomputing the historical decoder states.
\emph{Zero KV} replaces the complete key and value tensors with zeros of the
same shape. \emph{Random KV} samples independent Gaussian directions for the
keys and values and rescales each layer--head--token vector to match the norm of
its corresponding historical vector.

The three hard-insertion controls therefore share the same selected page,
number of inserted tokens, positional layout, visible attention span, query,
and benchmark-specific answer scorer, differing only in the inserted KV
content. Side-KV exposes the same complete historical page through an isolated
query-local Recall trajectory and releases the associated decoder state after
answering. When no historical page is causally eligible, all integration
variants use the same recent-context fallback. Table~\ref{tab:ablation-sidekv}
therefore compares full-page insertion through the Native-cache interface with
full-page query-local access through Side-KV under matched retrieved evidence
and page capacity.

\subsection{Efficiency Measurement Protocol}
\label{app:efficiency}

\paragraph{Timing setup.}
We measure ten video prefixes containing 26 supplied queries, which
together represent 1046.5\,s of video per repetition, under 1- and 2-FPS input
sampling. Each configuration runs on one NVIDIA A100-SXM4-80GB GPU with one
video stream at a time and four CPU compute threads. Native and FlashBack are
evaluated sequentially on the same physical GPU. Each system first processes
the first video once as an untimed warm-up. We then run two complete
repetitions, reversing the Native--FlashBack execution order in the second
repetition. CUDA is synchronized immediately before and after each timed
stream. Bars pool the two repetitions, while whiskers indicate the minimum and
maximum of their separately aggregated measurements.

\paragraph{What is included.}
Wall-clock timing covers causal video decoding and frame sampling, the
backbone's visual and language computation, and all scheduled query forwards.
For FlashBack, it additionally includes online memory construction,
temporal-evidence routing, candidate prefiltering, query--history Q--K
reranking, the Side-KV Recall forward when invoked, and candidate-answer logit
computation. Memory is maintained even for queries routed to Native. The Native
systems retain their original execution behavior; in particular, Mage-VL
buffers sampled frames and performs its native four-frame current-window
inference at query time rather than encoding every sampled frame. Model loading,
warm-up, diagnostic hashing, logging, output serialization, and lightweight
post-forward result formatting are outside the timed region. The reported time
therefore reflects the model-side online processing path evaluated in
Figure~\ref{fig:efficiency}.

\paragraph{Metrics.}
Let \(T_{r,v}\) denote the processing wall time, \(D_v\) the represented video
duration, and \(N_{r,v}\) the number of consumed sampled frames for video \(v\)
in repetition \(r\). Figure~\ref{fig:efficiency} reports processing cost as
\begin{equation}
60
\frac{\sum_{r,v} T_{r,v}}
     {\sum_{r,v} D_v},
\end{equation}
measured in seconds per minute of represented video, and processing throughput
as
\begin{equation}
\frac{\sum_{r,v} N_{r,v}}
     {\sum_{r,v} T_{r,v}},
\end{equation}
measured in frames per wall-clock second. Both metrics are computed from the
same unpaced workload. FlashBack uses the same representation-balanced router
and \(K=1\) setting as in the main experiments, rather than forcing Recall for
every query. In each repetition, 15 of the 26 queries invoke Recall for
StreamingVLM and 12 of 26 for Mage-VL. Throughput characterizes aggregate
processing capacity relative to the input sampling rate and is distinct from
per-query response latency.

\subsection{Memory-Bank Growth and Computational Complexity}
\label{app:memory_complexity}

\paragraph{Archived representations and growth.}
Historical evidence is retained as completed causal units whose realization
follows the backbone. StreamingVLM uses video chunks of up to four seconds,
whereas Mage-VL uses units of up to four sampled frames. For each StreamingVLM
unit, FlashBack retains the evicted key--value states at the retrieval layers,
the projected visual representations required for later page construction, and
a compact semantic representation used by the prefilter. Full-layer page KV is
materialized on demand for retrieved pages rather than stored for every
historical unit. Mage-VL retains the token IDs, decoder-input embeddings,
visual-token positions, and compact representations required for candidate
construction, while exact decoder keys are likewise materialized for
shortlisted pages when needed.

The reported configuration does not impose a capacity limit or evict completed
historical units from \(H_t\). The persistent Native streaming state remains
bounded by the underlying backbone, while the auxiliary historical archive
grows with the number and size of retained units. Under fixed sampling and page
settings, this storage therefore grows approximately linearly with observed
video duration; previously materialized page caches may contribute additional
storage.

\paragraph{Query-time computation.}
For a Recall query, causal eligibility checking and semantic prefiltering
traverse the archive and score the eligible units in \(\mathcal E_t\). In the
reported configuration, the candidate set \(\mathcal C_t\) is formed by taking
the highest-ranked unit for each query hypothesis and removing duplicates.
Decoder-level Q--K reranking, together with any required candidate-page KV
materialization, is then restricted to \(\mathcal C_t\), after which the
selected page is exposed through the Recall trajectory.

Accordingly, semantic prefiltering scales with the number of eligible
historical units \(|\mathcal E_t|\), whereas decoder-level reranking operates
only on the much smaller candidate set \(|\mathcal C_t|\). Archive storage
scales with the retained history, while the more expensive candidate-level
decoder computation is restricted to the shortlisted pages.

\subsection{Additional Visualization Study}
\label{app:qualitative}
\subsubsection{Visualization Examples}
We provide additional visualization cases to complement the examples shown in
the main text. Each case visualizes the retrieved historical page, its temporal
location in the observed stream, and the corresponding Native and Recall
answers. The examples illustrate different outcomes of historical Recall under
different temporal-evidence requirements.
\vspace{2mm}
\begin{figure*}[h]
    \centering
    \includegraphics[width=\textwidth]
    {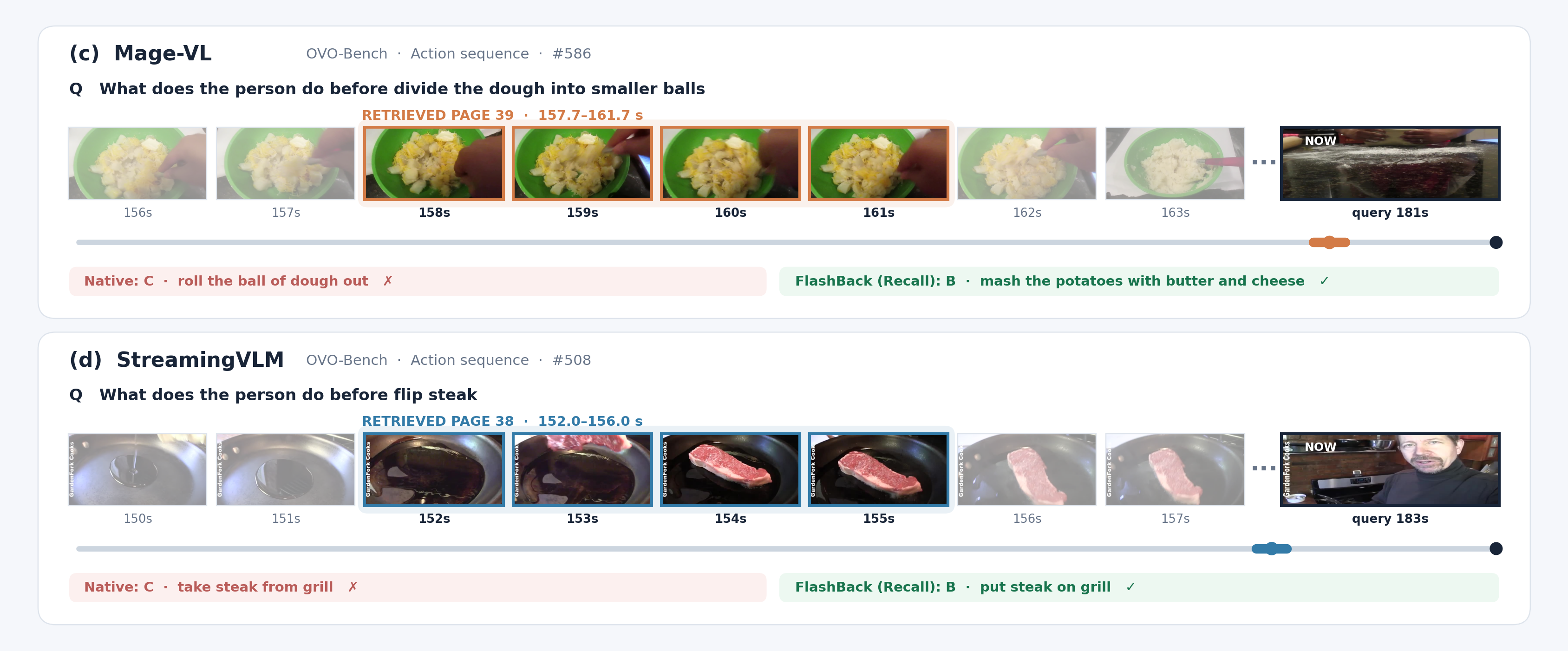}
    \caption{Additional FlashBack visualization cases.}
    \label{fig:qualitative_success_additional}
\end{figure*}

Figure~\ref{fig:qualitative_success_additional} shows two additional
action-sequence examples from Mage-VL and StreamingVLM. In
Figure~\ref{fig:qualitative_success_additional}(c), the query asks what occurs
before the dough is divided into smaller balls. FlashBack retrieves the earlier
preparation segment at 157.7--161.7\,s and identifies the relevant action,
whereas the Native trajectory selects an unrelated event. In
Figure~\ref{fig:qualitative_success_additional}(d), the retrieved
152--156\,s segment contains the action of placing the steak on the grill,
providing the evidence required to answer what occurs before the later flipping
action. In both cases, Recall makes the relevant earlier event available again
at query time when it is no longer contained in the immediate stream context.
\vspace{2mm}
\begin{figure*}[h]
    \centering
    \includegraphics[width=\textwidth]
    {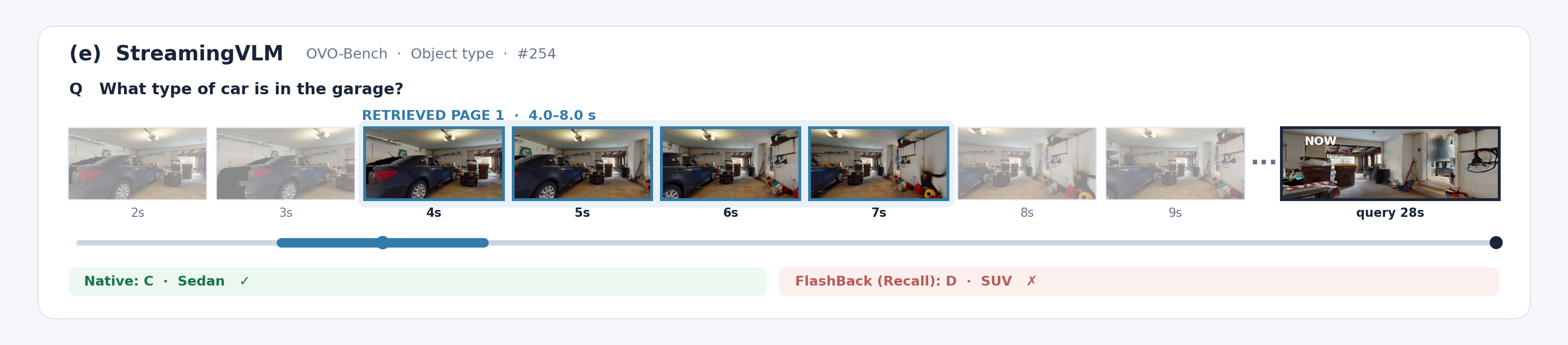}
    \caption{A contrasting FlashBack visualization case.}
    \label{fig:qualitative_failure}
\end{figure*}

Figure~\ref{fig:qualitative_failure} shows a contrasting case in which the
Native trajectory correctly identifies the car as a \emph{Sedan}, whereas
FlashBack predicts \emph{SUV} after Recall. The retrieved 4--8\,s page does
contain the queried vehicle, so the error is not caused by the absence of
relevant historical evidence. Instead, the failure occurs when the Recall
trajectory uses this evidence for the finer-grained vehicle-type decision.

This case distinguishes retrieving relevant history from benefiting from it
during answering. A historical page can be temporally and semantically relevant
without necessarily improving the downstream prediction. Together with the
successful cases above, this separates memory access from evidence utilization:
Recall can make relevant historical evidence available, while the final
prediction still depends on how the frozen backbone interprets that evidence
during answering.

\section{Complete Task-Level Results}
\label{app:detailed-results}
\subsection{Complete OVO-Bench Results}
Table~\ref{tab:ovo-complete} reports the complete task-level results on
OVO-Bench.
\begin{table*}[h]
\centering
\captionsetup{
    font=footnotesize,
    labelfont=normalfont,
    textfont=bf,
    skip=3pt
}
\caption{\textbf{Detailed OVO-Bench results of FlashBack.}}
\label{tab:ovo-complete}

\vspace{-1mm}
\setlength{\tabcolsep}{5.0pt}
\renewcommand{\arraystretch}{1.06}

\resizebox{1\linewidth}{!}{%
\begin{tabular}{@{}l|
cccccc|c|
ccc|c|
ccc|c|
c@{}}
\toprule

\multirow{2}{*}{\textbf{Method}}
& \multicolumn{7}{c|}{\textbf{Realtime}}
& \multicolumn{4}{c|}{\textbf{Backward}}
& \multicolumn{4}{c|}{\textbf{Forward}}
& \multirow{2}{*}{\textbf{Overall}} \\

\cmidrule(lr){2-8}
\cmidrule(lr){9-12}
\cmidrule(lr){13-16}

& \textbf{OCR}
& \textbf{ACR}
& \textbf{ATR}
& \textbf{STU}
& \textbf{FPD}
& \textbf{OJR}
& \textbf{Avg.}
& \textbf{EPM}
& \textbf{ASI}
& \textbf{HLD}
& \textbf{Avg.}
& \textbf{REC}
& \textbf{SSR}
& \textbf{CRR}
& \textbf{Avg.}
& \\
\midrule

StreamingVLM Native-7B
& \textbf{91.28} & \textbf{65.14} & \textbf{75.00} & 48.88 & \textbf{70.30} & \textbf{79.35}
& \textbf{71.66}
& \textbf{54.88} & \textbf{56.76} & \textbf{27.96}
& \textbf{46.53}
& 19.34 & \textbf{42.77} & 41.67
& 34.59
& 50.93 \\

\rowcolor{flashblue}
\textbf{StreamingVLM + FlashBack-7B}
& 90.60 & 64.22 & \textbf{75.00} & \textbf{49.44} & \textbf{70.30} & 75.54
& 70.85
& \textbf{54.88} & \textbf{56.76} & 23.12
& 44.92
& \textbf{20.06} & \textbf{42.77} & \textbf{60.83}
& \textbf{41.22}
& \textbf{52.33} \\
\midrule
Mage-VL Native-4B
& \textbf{96.64} & 80.73 & \textbf{85.34} & \textbf{74.16} & \textbf{72.28} & \textbf{81.52}
& \textbf{81.78}
& 52.53 & 58.78 & \textbf{32.26}
& 47.86
& 19.48 & 71.86 & 44.17
& 45.17
& 58.27 \\

\rowcolor{flashblue}
\textbf{Mage-VL + FlashBack-4B}
& \textbf{96.64} & \textbf{81.65} & \textbf{85.34} & 73.03 & \textbf{72.28} & 78.26
& 81.20
& \textbf{53.20} & \textbf{64.86} & 31.72
& \textbf{49.93}
& \textbf{29.80} & \textbf{71.86} & \textbf{54.17}
& \textbf{51.94}
& \textbf{61.02} \\

\bottomrule
\end{tabular}%
}

\end{table*}
\subsection{Detailed StreamingBench Results}
Tables~\ref{tab:main-streamingbench-rt} and ~\ref{tab:main-streamingbench-full} report the detailed task-level
results on StreamingBench.

\vspace{2mm}
\begin{table*}[h]
\captionsetup{
    font=footnotesize,
    labelfont=normalfont,
    textfont=bf,
    skip=3pt
}
\centering
\caption{\textbf{Real-Time Visual Understanding results on StreamingBench.}}
\label{tab:main-streamingbench-rt}

\vspace{-1mm}
\setlength{\tabcolsep}{5.0pt}
\renewcommand{\arraystretch}{1.06}

\resizebox{0.95\linewidth}{!}{%
\begin{tabular}{@{}l|
cccccccccc|c@{}}
\toprule

\multirow{2}{*}{\textbf{Method}}
& \multicolumn{11}{c}{\textbf{Real-Time Visual Understanding}} \\
\cmidrule(lr){2-12}

& \textbf{OP}
& \textbf{CR}
& \textbf{CS}
& \textbf{ATP}
& \textbf{EU}
& \textbf{TR}
& \textbf{PR}
& \textbf{SU}
& \textbf{ACP}
& \textbf{CT}
& \textbf{Avg.} \\
\midrule

\multicolumn{12}{@{}l}{\color{traintext}\textit{Training-based methods}} \\

VideoLLM-online
& 39.07 & 40.06 & 34.49 & 31.05 & 45.96
& 32.40 & 31.48 & 34.16 & 42.49 & 27.89
& 35.99 \\

EventMemAgent-8B
& 83.92 & 76.56 & 87.70 & 85.29 & 77.02
& 79.44 & 77.78 & 68.29 & 72.24 & 48.70
& 77.00 \\

SelectStream-Qwen3-8B
& 86.30 & 87.40 & 86.00 & 87.10 & \textbf{84.70}
& 84.60 & \textbf{85.80} & 75.00 & 78.20 & \textbf{71.60}
& \textbf{82.67} \\

\midrule
\multicolumn{12}{@{}l}{\color{traintext}\textit{Training-free methods}} \\

StreamRAG-ViSpeak-7B
& 83.37 & \textbf{90.63} & 86.43 & 83.99 & 78.26
& 85.67 & 77.78 & 73.17 & 77.90 & 31.10
& 78.12 \\

HERMES-7B 
& 83.65 & 81.25 & 88.01 & 87.46 & 76.73
& 86.60 & 82.41 & 76.02 & 73.94 & 46.63
& 79.44 \\

\midrule
\multicolumn{12}{@{}l}{\color{traintext}\textit{Ours (training-free)}} \\

StreamingVLM Native-7B
& 84.55 & 79.69 & 87.70 & 88.75 & 80.38
& 89.72 & 80.56 & 74.80 & 74.72 & 27.66
& 78.82 \\

\rowcolor{flashblue}
\textbf{StreamingVLM + FlashBack-7B}
& 84.55 & 81.25 & 88.33 & 88.75 & 80.38
& 90.03 & 80.56 & 74.80 & 74.72 & 30.32
& 79.22 \\

Mage-VL Native-4B
& \textbf{90.24} & 67.97 & \textbf{90.22} & \textbf{89.71} & 76.58
& \textbf{90.34} & 78.70 & \textbf{80.49} & 80.68 & 40.43
& 81.63 \\

\rowcolor{flashblue}
\textbf{Mage-VL + FlashBack-4B}
& \textbf{90.24} & 71.88 & \textbf{90.22} & \textbf{89.71} & 81.01
& \textbf{90.34} & 78.70 & \textbf{80.49} & \textbf{80.97} & 46.28
& 82.59 \\

\bottomrule
\end{tabular}%
}

\vspace{4mm}
\end{table*}

\begin{table*}[h]
\centering
\captionsetup{
    font=footnotesize,
    labelfont=normalfont,
    textfont=bf,
    skip=3pt
}
\caption{\textbf{Detailed StreamingBench results of FlashBack.}}
\label{tab:main-streamingbench-full}

\vspace{-1mm}
\setlength{\tabcolsep}{5.0pt}
\renewcommand{\arraystretch}{1.06}

\resizebox{0.95\linewidth}{!}{%
\begin{tabular}{@{}l|
c|
cccc|c|
ccc|c|
c@{}}
\toprule

\multirow{2}{*}{\textbf{Method}}
& \multirow{2}{*}{\textbf{RT Avg.}}
& \multicolumn{5}{c|}{\textbf{Omni-Source Understanding}}
& \multicolumn{4}{c|}{\textbf{Contextual Understanding}}
& \multirow{2}{*}{\textbf{Overall}} \\

\cmidrule(lr){3-7}
\cmidrule(lr){8-11}

&
&
\textbf{ER}
& \textbf{SCU}
& \textbf{SD}
& \textbf{MA}
& \textbf{Avg.}
& \textbf{ACU}
& \textbf{MCU}
& \textbf{SQA}
& \textbf{Avg.}
& \\

\midrule

StreamingVLM Native-7B
& 78.82
& \textbf{50.00} & \textbf{40.80} & 51.20 & 69.60 & 52.90
& 52.40 & \textbf{41.60} & \textbf{50.40} & 48.13
& 67.30 \\

\rowcolor{flashblue}
\textbf{StreamingVLM + FlashBack-7B}
& \textbf{79.22}
& 49.60 & 38.80 & \textbf{52.00} & \textbf{72.80} & \textbf{53.30}
& \textbf{53.60} & 41.20 & 50.00 & \textbf{48.27}
& \textbf{67.66} \\
\midrule
Mage-VL Native-4B
& 81.63
& \textbf{56.00}
& \textbf{49.60}
& \textbf{64.40}
& \textbf{90.80}
& \textbf{65.20}
& \textbf{67.20}
& \textbf{64.80}
& \textbf{54.40}
& \textbf{62.13}
& 74.32 \\

\rowcolor{flashblue}
\textbf{Mage-VL + FlashBack-4B}
& \textbf{82.59}
& \textbf{56.00}
& \textbf{49.60}
& \textbf{64.40}
& 90.40
& 65.10
& \textbf{67.20}
& 61.60
& \textbf{54.40}
& 61.07
& \textbf{74.67} \\

\bottomrule
\end{tabular}%
}

\end{table*}
\end{document}